\def\year{2022}\relax
\documentclass[letterpaper]{article} 
\usepackage{aaai22}  
\usepackage{times}  
\usepackage{helvet}  
\usepackage{courier}  
\usepackage[hyphens]{url}  
\usepackage{graphicx} 
\usepackage{natbib}  
\usepackage{caption} 
\DeclareCaptionStyle{ruled}{labelfont=normalfont,labelsep=colon,strut=off} 
\usepackage{algorithm}
\usepackage{algorithmic}

\usepackage{newfloat}
\usepackage{listings}
\floatstyle{ruled}
\newfloat{listing}{tb}{lst}{}
\floatname{listing}{Listing}

\usepackage{amsmath}
\usepackage{array}
\usepackage{booktabs}
\usepackage{enumitem}
\usepackage{subcaption}
\usepackage{pifont}
\usepackage{multirow}
\usepackage[frozencache=true,cachedir=.]{minted}
\usepackage{color}

\usepackage{xspace}
\newcommand{\dataset}[1]{\textit{#1}\xspace}
\newcommand{\teach}{\dataset{TEACh}}
\newcommand{\teachfull}{\dataset{\textbf{T}ask-driven \textbf{E}mbodied \textbf{A}gents that \textbf{C}hat}}

\newcommand{\commander}{\textit{Commander}}
\newcommand{\follower}{\textit{Follower}}

\newcommand{\task}[1]{\textsc{#1}}
\newcommand{\action}[1]{\texttt{#1}}
\newcommand{\object}[1]{\texttt{#1}}
\newcommand{\cmark}{\raisebox{0pt}{\ding{51}}}
\newcommand{\xmark}{\raisebox{0pt}{\ding{55}}}
\newcommand{\na}{-}

\newcolumntype{L}[1]{>{\raggedright\let\newline\\\arraybackslash\hspace{0pt}}m{#1}}
\newcolumntype{C}[1]{>{\centering\let\newline\\\arraybackslash\hspace{0pt}}m{#1}}
\newcolumntype{R}[1]{>{\raggedleft\let\newline\\\arraybackslash\hspace{0pt}}m{#1}}

\mathchardef\mhyphen="2D

\newcommand\blfootnote[1]{%
  \begingroup
  \renewcommand\thefootnote{}\footnote{#1}%
  \addtocounter{footnote}{-1}%
  \endgroup
}

\title{TEACh: Task-driven Embodied Agents that Chat}
\author {
    Aishwarya Padmakumar\textsuperscript{*} \textsuperscript{\rm 1},
    Jesse Thomason\textsuperscript{*} \blfootnote{ Authors contributed equally} \textsuperscript{\rm 1} \textsuperscript{\rm 2},
    Ayush Shrivastava\textsuperscript{\rm 3},
    Patrick Lange\textsuperscript{\rm 1},
    Anjali Narayan-Chen\textsuperscript{\rm 1},
    Spandana Gella\textsuperscript{\rm 1},
    Robinson Piramuthu\textsuperscript{\rm 1},
    Gokhan Tur\textsuperscript{\rm 1},
    Dilek-Hakkani Tur\textsuperscript{\rm 1}
}
\affiliations {
    \textsuperscript{\rm 1} Amazon Alexa AI \\
    \textsuperscript{\rm 2} USC Viterbi Department of Computer Science, University of Southern California \\
    \textsuperscript{\rm 3} Department of Electrical Engineering And Computer Science, University of Michigan \\
    padmakua@amazon.com, jessedt@amazon.com, ayshrv@umich.edu, patlange@amazon.com, naraanja@amazon.com, sgella@amazon.com, robinpir@amazon.com, gokhatur@amazon.com, hakkanit@amazon.com
}

\begin{document}

\maketitle

\begin{abstract}
Robots operating in human spaces must be able to engage in natural language interaction, both understanding and executing instructions, and using conversation to resolve ambiguity and correct mistakes.
To study this, we introduce \teach, a dataset of over 3,000 human--human, interactive dialogues to complete household tasks in simulation.
A \commander\ with access to oracle information about a task communicates in natural language with a \follower. 
The \follower\ navigates through and interacts with the environment to complete tasks varying in complexity from \task{Make Coffee} to \task{Prepare Breakfast}, asking questions and getting additional information from the \commander.
We propose three benchmarks using \teach\ to study embodied intelligence challenges, and we evaluate initial models' abilities in dialogue understanding, language grounding, and task execution.
\end{abstract}


\section{Introduction}

\begin{figure}[t]
    \centering
    \includegraphics[width=.85\columnwidth]{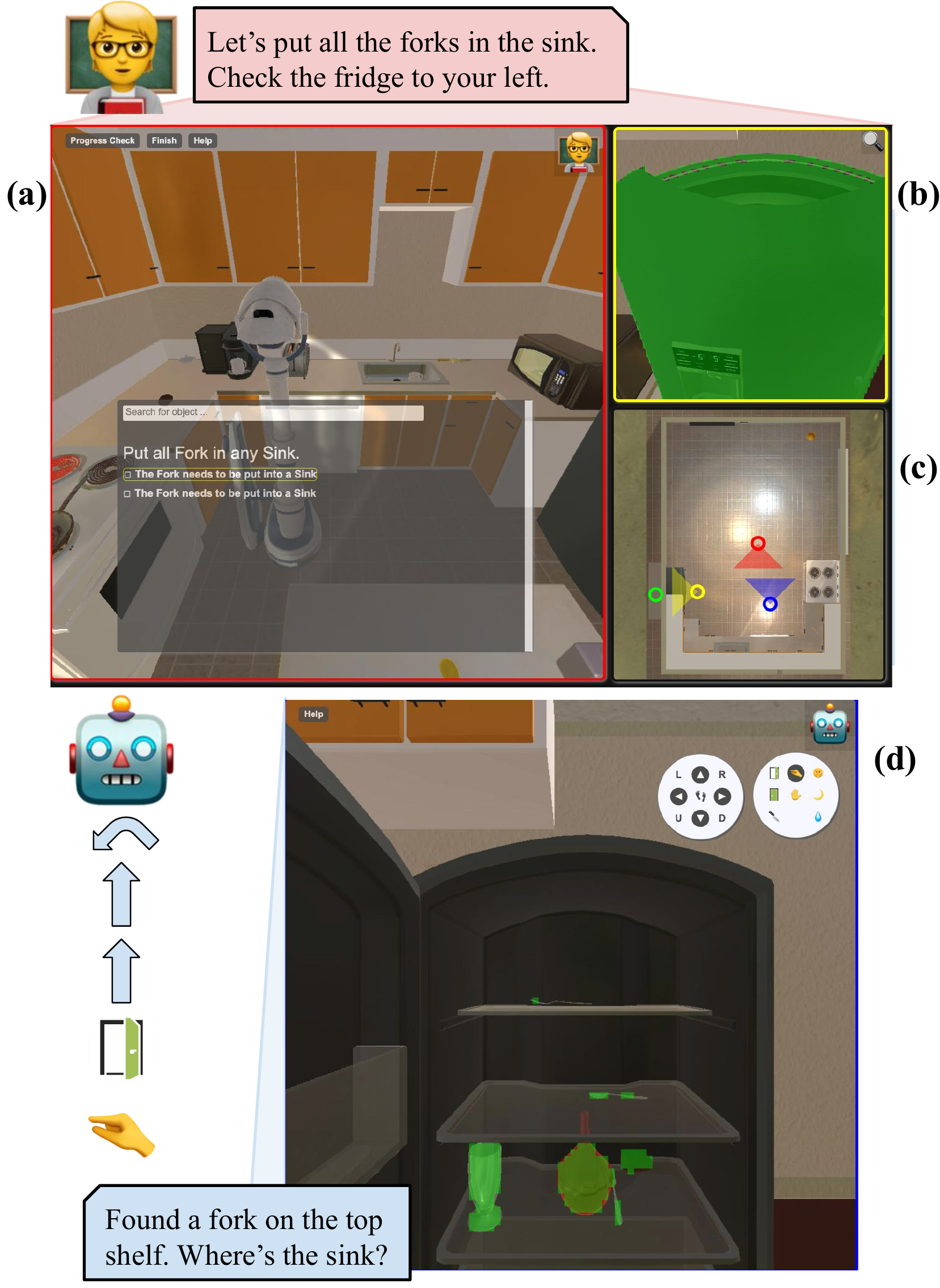}
    \caption{
    The \commander\ has oracle task details (a), object locations (b), a map (c), and egocentric views from both agents.
    The \follower\ carries out the task and asks questions (d).
    The agents can only communicate via language.
    }
    \label{fig:teaser}
\end{figure}

Many benchmarks for translating visual observations and an initial language instruction to actions 
assume no further language communication~\cite{anderson:cvpr18,shridhar:cvpr20}.
However, obtaining clarification via simulated interactions~\cite{chi:aaai20,nguyen:emnlp19} or learning from human-human dialogue~\cite{thomason:corl19,suhr:emnlp19} can improve embodied navigation.
We hypothesize that dialogue has even more to offer for object-centric, hierarchical tasks.

\begin{table*}[t]
\centering 
\tabcolsep 2.5pt
\begin{tabular}{lcccccc}
     \bf Dataset & \multicolumn{2}{c}{\bf --- Object ---} & \multicolumn{3}{c}{\bf --- Language ---} & \bf Demonstrations \\    
     & Interaction & State Changes & Conversational & \# Sessions & Freeform & \\
     \toprule
     R2R~\cite{anderson:cvpr18} & \xmark & \xmark & \xmark & \na & \na & Planner \\
     CHAI~\cite{misra:emnlp18} & \cmark & \cmark & \xmark & \na & \na & Human \\
     CVDN~\cite{thomason:corl19} & \xmark & \xmark & \cmark & 2050 & \xmark & Human \\
     CerealBar~\cite{suhr:emnlp19} & \cmark & \xmark & \cmark & 1202 & \xmark & Human \\
     MDC~(Narayan-Chen et al. \citeyear{narayan-chen:acl19}) & \cmark & \xmark & \cmark & 509 & \cmark & Human \\
     ALFRED~\cite{shridhar:cvpr20} & \cmark & \cmark & \xmark & \na & \na & Planner \\
     III~\cite{imitating_interactive_intelligence} & \cmark & \xmark & \xmark & - & \na & Human \\
     \midrule
     \teach & \cmark & \cmark & \cmark & 3215 & \cmark & Human \\
     \bottomrule
\end{tabular}
\caption{\teach\ is the first dataset where human-human, conversational dialogues were used to perform tasks involving object interaction, such as picking up a knife, and state changes, such as slicing bread, in a visual simulation environment.
\teach\ task demonstrations are created by the human \follower, who engages in a free-form, rather than turn-taking, dialogue with the human \commander.
Compared to past dialogue datasets for visual tasks, \teach\ contains many more individual dialogues.
}
    \label{tab:benchmark_qualitative_comparison}
\end{table*}

We introduce \teachfull\ (\teach)\footnote{https://github.com/alexa/teach} to study how agents can learn to ground natural language~\cite{harnad:90,egl} to the visual world and actions, while considering long-term and intermediate goals, and using dialogue to communicate. 
\teach\ contains over 3,000 human--human sessions interleaving utterances and environment actions where a \commander\ with oracle task and world knowledge and a \follower\ with the ability to interact with the world communicate in written English to complete household chores (Figure~\ref{fig:teaser}).


\teach\ dialogues are unconstrained, not turn-based, yielding variation in instruction granularity, completeness, relevance, and overlap.
Utterances include coreference with previously mentioned entities, past actions, and locations.
Because \teach\ sessions are human, rather than planner-based~\cite{pddl}, \follower\ trajectories include mistakes and corresponding, language-guided correction.

We propose three benchmarks based on \teach\ sessions to study the ability of learned models to achieve aspects of embodied intelligence: Execution from Dialog History (EDH), Trajectory from Dialog (TfD) and Two-Agent Task Completion (TATC).
We also demonstrate baseline performance on these benchmarks.
To model the \follower\ agent for the EDH and TfD benchmarks, we build on the Episodic Transformer (E.T.) model~\cite{pashevich:arxiv21}
as a baseline.
When modeling both agents for end-to-end task completion, we demonstrate the difficulty of engineering rule-based solvers.

The main contributions of this work are:
\begin{itemize}
\setlength\topsep{0em}
\setlength\itemsep{0em}
    \item \teach, a dataset of over 3000 human-human dialogs simulating the experience of a user interacting with their robot to complete tasks in their home, that interleaves dialogue messages with actions taken in the environment. 
    \item An extensible task definition framework (\S\ref{sec:dataset}) that can be used to define and check completion status for a wide range of tasks in a simulated environment. 
    \item Three benchmarks based on \teach\ sessions and experiments demonstrating initial models for each.
\end{itemize}


\section{Related Work}


Table \ref{tab:benchmark_qualitative_comparison} situates \teach\ with respect to other datasets involving natural language instructions for visual task completion.

\paragraph{Vision $\&$ Language Navigation (VLN)} tasks agents with taking in language instructions and a visual observation to produce an action, such as turning or moving forward, to receive a new visual observation.
VLN benchmarks have evolved from the use of symbolic environment representations~\cite{macmahon:aaai06,chen:aaai11,mei:aaai16} to photorealistic indoor~\cite{anderson:cvpr18} and outdoor environments~\cite{chen:cvpr19}, as well as the prediction of continuous control~\cite{blukis:corl18}.
\teach\ goes beyond navigation to object interactions for task completion, and beyond single instructions to dialogue.

\paragraph{Vision $\&$ Language Task Completion} involves actions beyond navigation.
Models have evolved from individual rule-based or learned components for language understanding, perception and action execution~\cite{matuszek:2013,kollar:jair13}, to end-to-end models in fully observable blocks worlds~\cite{bisk:aaai18,misra:emnlp18}.
More complex tasks involve partially observable worlds~\cite{kim:emnlp20} and object state changes~\cite{misra:emnlp18,puig:cvpr18,shridhar:cvpr20}.
Some works use a planner to generate ideal demonstrations that are then labeled, while others first gather instructions and gather human demonstrations~\cite{misra:emnlp18,shah:arxiv21,imitating_interactive_intelligence}.
In \teach , human instructions and demonstrations are gathered simultaneously.
\begin{figure*}[t]
    \centering
    \includegraphics[width=.9\textwidth]{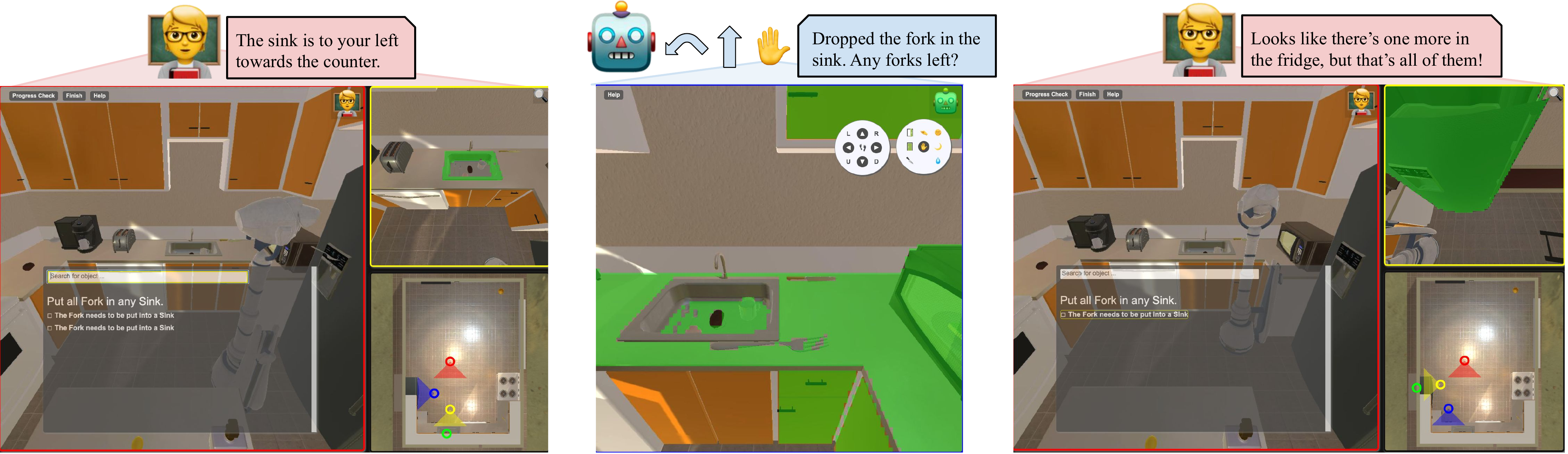}
    \caption{To collect \teach, the \commander\ knows the task to be completed and can query the simulator for object locations.
    Searched items are highlighted in green for the \commander; highlights blink to enable seeing the underlying true scene colors.
    The \commander\ has a topdown map of the scene, with the current camera position shown in red, the \follower\ position shown in blue, and the object search camera position shown in yellow.
    The \follower\ moves around in the environment and interacts with objects, such as placing a fork (middle).
    Target objects for each interaction action are highlighted.
    }
    \label{fig:interface}
\end{figure*}

\paragraph{Vision $\&$ Dialogue Navigation and Task Completion}
Agents that engage in dialogue instead of simply following natural language instructions can be learned by combining individual rule-based or learned components~\cite{tellex:rss14,arumugam:ar18,thomason:jair20}.  
Simulated clarification can also improve end-to-end VLN models~\cite{chi:aaai20,nguyen:emnlp19}.
Models are also able to take advantage of conversational history in human-human dialogues to perform better navigation~\cite{thomason:corl19,babywalk}, learn agent-agent policies for navigating and speaking~\cite{rmm,shrivastava:arxiv21}, and deploy individual agent policies for human-in-the-loop evaluation~\cite{suhr:emnlp19}.
However, such models and underlying datasets are limited to navigation actions and turn-taking conversation.
In contrast, \teach\ involves \follower\ navigation and object interaction, as well as freeform dialogue acts with the \commander.
The Minecraft Dialogue Corpus (MDC)~\cite{narayan-chen:acl19} gives full dialogues between two humans for assembly tasks.
MDC is similar in spirit to \teach; we introduce a larger action space and resulting object state changes, such as slicing and toasting bread, as well as collecting many more human-human dialogues.

\section{The \teach\ Dataset}
\label{sec:dataset}

We collect 3,047 human--human \textit{gameplay sessions} for completing household tasks in the AI2-THOR simulator~\cite{kolve:17}. 
Each session includes an initial environment state, \commander\ actions to access oracle information, utterances between the \commander\ and \follower, movement actions, and object interactions taken by the \follower.
Figure~\ref{fig:interface} gives an overview of the annotation interface.


\subsection{Household Tasks}

\begin{figure*}[t]
    \centering
    \includegraphics[width=.9\textwidth]{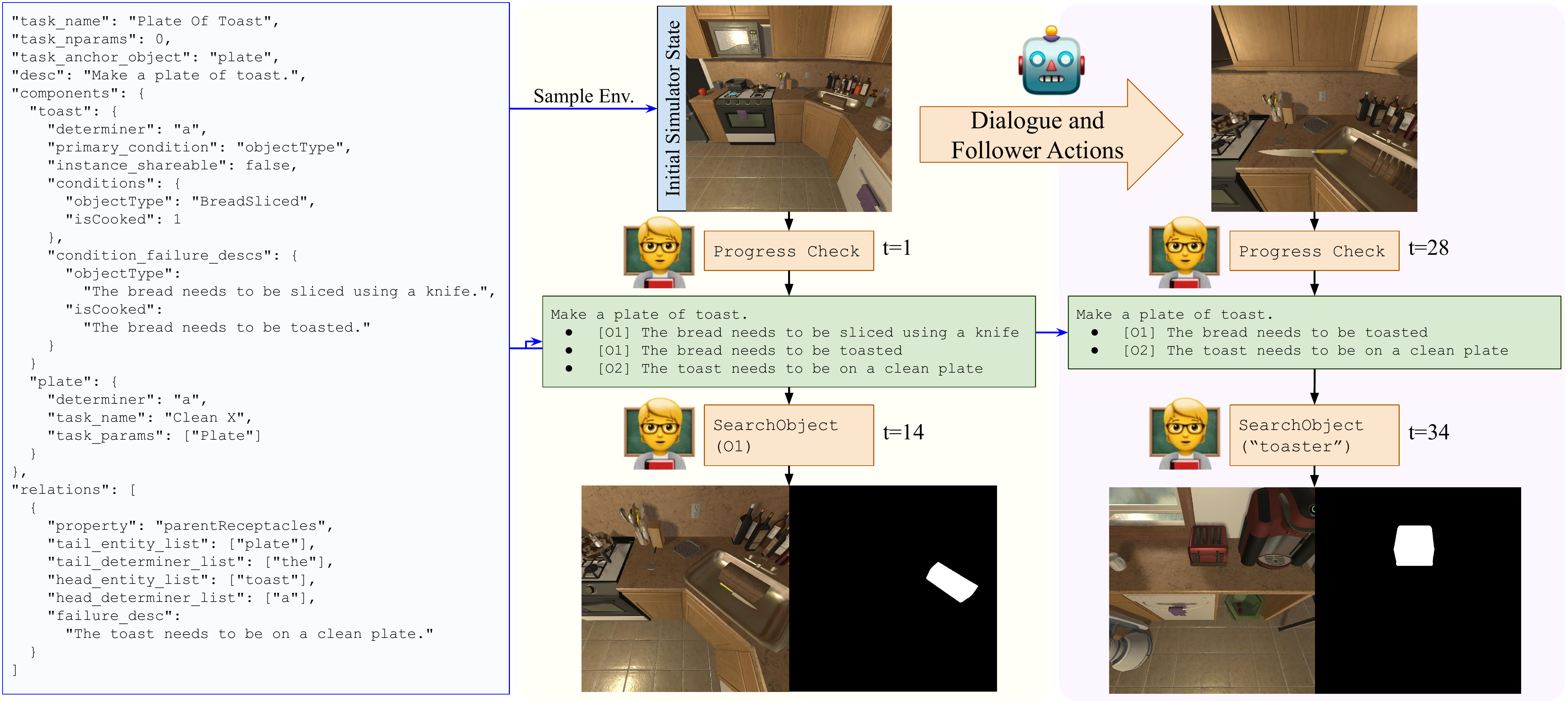}
    \caption{An example task definition from the \teach\ task definition language (left) and how it informs the initial simulator state and the \commander\ \action{Progress Check} action.
    The \commander\ can \action{SearchObject} with a string query (right) or object instance (center) returned by the \action{Progress Check} task status, yielding a camera view, segmentation mask, and location.
    }
    \label{fig:tdl_example}
\end{figure*}

We design a \textit{task definition language} (TDL) to define household tasks in terms of object properties to satisfy, and implement a framework over AI2-THOR that evaluates these criteria.
For example, for a task to make coffee, we consider the environment to be in a successful state if there is a mug in the environment that is clean and filled with coffee. 

Parameterized tasks such as \task{Put All X On Y} enable task variation.
Parameters can be object classes, such as putting all \object{forks} on a \object{countertop}, or predefined abstract hypernyms, for example putting all \object{silverware}---forks, spoons, and knives---on the counter.
\teach\ task definitions are also hierarchical. 
For example, \task{Prepare Breakfast} contains the subtasks \task{Make Coffee} and \task{Make Plate of Toast}.
We incorporate determiners such as ``a'', ``all'' and numbers such as 2 to enable easy definition of a wide range of tasks, such as \task{N Slices of X in Y}.
The \teach\ TDL includes template-based language prompts to describe tasks and subtasks to \commander{}s (Figure~\ref{fig:tdl_example}).

\subsection{Gameplay Session Collection}
\label{ssec:data_collection_setup}

Annotators first completed a tutorial task demonstrating the interface to vet their understanding.
For each session, two vetted crowdworkers were paired using a web interface and assigned to the \commander\ and \follower\ roles (Figure~\ref{fig:interface}).
The \commander\ is shown the task to be completed and the steps needed to achieve this given the current state of the environment, using template-based language prompts, none of which are accessible to the \follower.
The \commander\ can additionally search for the location of objects, either by string name, such as ``sink'', or by clicking a task-relevant object in the display (Figure~\ref{fig:tdl_example}).
The \commander\ and \follower\ must use text chat to communicate the parameters of the task and clarify object locations. 
Only the \follower\ can interact with objects in the environment.

We obtained initial states for each parameterized task by randomizing AI2-THOR environments and retaining those that satisfied preconditions such as task-relevant objects being present and reachable.
For each session, we store the initial simulator state $S_i$, the sequence of actions $A = (a_1, a_2, \ldots )$ taken, and the final simulator state $S_f$.
\teach\ \follower\ actions are \action{Forward}, \action{Backward}, \action{Turn Left}, \action{Turn Right}, \action{Look Up}, \action{Look Down}, \action{Strafe Left}, \action{Strafe Right}, \action{Pickup}, \action{Place}, \action{Open}, \action{Close}, \action{ToggleOn}, \action{ToggleOff}, \action{Slice}, and \action{Pour}.
Navigation actions move the agent in discrete steps.
Object manipulation expects the agent to specify an object via a relative coordinate $(x, y)$ on \follower\ egocentric frame.
The \teach\ wrapper on the AI2-THOR simulator examines the ground truth segmentation mask of the agent's egocentric image, selects an object in a 10x10 pixel patch around the coordinate if the desired action can be performed on it, and executes the action in AI2-THOR.
The \commander\ can execute a \action{Progress Check} and \action{SearchObject} actions, demonstrated in Figure~\ref{fig:tdl_example}.
\teach\ \commander\ actions also allow navigation, but the \commander\ is a disembodied camera.

\begin{table*}[t]
    \centering
    \begin{tabular}{lrrrrrr}
         & \multicolumn{1}{c}{Parameter} & \multicolumn{1}{c}{Unique} & \multicolumn{1}{c}{Total} & \multicolumn{1}{c}{Utterances} & \multicolumn{1}{c}{\follower} & \multicolumn{1}{c}{All} \\
          & \multicolumn{1}{c}{Variants} & \multicolumn{1}{c}{Scenes} & \multicolumn{1}{c}{Sessions} & \multicolumn{1}{c}{per Session} & \multicolumn{1}{c}{Actions/Session} & \multicolumn{1}{c}{Actions/Session}  \\
\toprule
         \task{Water Plant} 
            & 1 & 10 & 176 & 6.37$\pm$\phantom{0}4.36 & 51.86$\pm$\phantom{0}30.71 & 67.93$\pm$\phantom{0}40.70 \\
         \task{Make Coffee} 
            & 1 & 30 & 308 & 7.75$\pm$\phantom{0}5.08 & 55.25$\pm$\phantom{0}33.61 & 72.29$\pm$\phantom{0}50.85 \\
         \task{Clean All X} 
            & 19 & 52 & 336 & 9.65$\pm$\phantom{0}7.03 & 74.06$\pm$\phantom{0}59.66 & 96.92$\pm$\phantom{0}71.31 \\
         \task{Put All X On Y} 
            & 209 & 92 & 344 & 8.66$\pm$\phantom{0}5.82 & 82.13$\pm$\phantom{0}66.39 & 103.53$\pm$\phantom{0}80.97 \\
         \task{Boil Potato} 
            & 1 & 26 & 202 & 10.65$\pm$\phantom{0}7.61 & 104.66$\pm$\phantom{0}79.50 & 130.13$\pm$\phantom{0}94.80 \\
         \task{Make Plate of Toast} 
            & 1 & 27 & 225 & 12.26$\pm$\phantom{0}8.51 & 108.30$\pm$\phantom{0}55.81 & 136.11$\pm$\phantom{0}70.73 \\
         \task{N Slices Of X In Y} 
            & 16 & 29 & 304 & 13.50$\pm$10.86 & 113.62$\pm$\phantom{0}94.25 & 146.23$\pm$113.96 \\
         \task{Put All X In One Y} 
            & 84 & 50 & 302 & 11.32$\pm$\phantom{0}7.03 & 115.74$\pm$\phantom{0}90.13 & 147.80$\pm$104.45 \\
         \task{N Cooked X Slices In Y} 
            & 10 & 30 & 240 & 14.94$\pm$\phantom{0}9.43 & 155.18$\pm$\phantom{0}75.17 & 189.26$\pm$\phantom{0}87.90 \\
         \task{Prepare Sandwich} 
            & 5 & 28 & 241 & 18.03$\pm$\phantom{0}9.96 & 195.93$\pm$\phantom{0}83.96 & 241.61$\pm$100.86 \\
         \task{Prepare Salad} 
            & 9 & 30 & 323 & 20.47$\pm$10.80 & 206.29$\pm$111.47 & 253.94$\pm$130.09 \\
         \task{Prepare Breakfast} 
            & 80 & 30 & 308 & 27.67$\pm$14.73 & 295.06$\pm$138.76 & 359.90$\pm$162.33 \\
         \midrule
         \bf \teach\ Overall & \bf 438 & \bf 109 & \bf 3320 & \textbf{13.67}$\pm$10.81 & \textbf{131.80}$\pm$109.68 & \textbf{164.65}$\pm$130.89 \\
         \bottomrule
    \end{tabular}
    \caption{The 12 tasks represented in \teach\ sessions vary in complexity.
    Tasks like \task{Put All X On Y} take object class parameters and can require more actions per session to finish.
    Composite tasks like \task{Prepare Salad} contain sub-tasks like \task{N Slices Of X In Y}.
    Per session data are averages with standard deviation across task types.
    }
    \label{tab:per_task_stats}
\end{table*}

\subsection{\teach\ Statistics}
\teach\ is comprised of 3,047 successful gameplay sessions, each of which can be replayed using the AI2-THOR simulator for model training, feature extraction, or model evaluation.
In total, 4,365 crowdsourced sessions were collected with a human-level success rate of 74.17\% (3320 sessions) and total cost of \$105k; more details in appendix.
Some successful sessions were not included in the final split used in benchmarks due to replay issues. 
\teach\ sessions span all 30 AI2-THOR kitchens, and include most of the 30 each AI2-THOR living rooms, bedrooms, and bathrooms.


Successful \teach\ sessions consist of over 45k utterances, with an average of 8.40 \commander\ and 5.25 \follower\ utterances per session. 
The average \commander\ utterance length is 5.70 tokens and the average \follower\ utterance length is 3.80 tokens.
The \teach\ data has a vocabulary size of 3,429 unique tokens.\footnote{Using the spaCy tokenizer: \url{https://pypi.org/project/spacy/}}
Table~\ref{tab:per_task_stats} summarizes such metrics across the 12 task types in \teach.
Simple tasks like \task{Make Coffee} require fewer dialogue acts and \follower\ actions on average than complex, composite tasks like \task{Prepare Breakfast} which subsume those simpler tasks.





\begin{figure*}[t]
    \centering
    \includegraphics[width=1\textwidth]{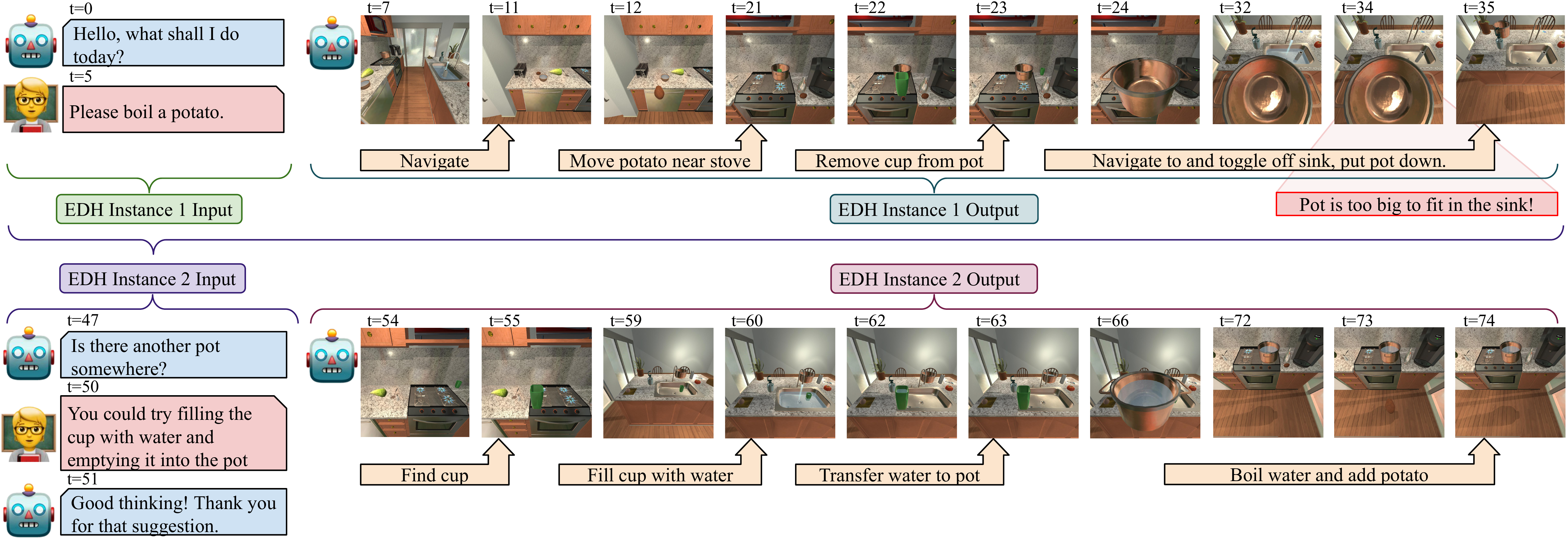}
    \caption{Two EDH instances are constructed from this real example from the \teach\ data.
    The first instance input contains only dialogue actions.
    After inference on the first instance, the agent is evaluated based on whether it moved the potato, pot, and the items cleared out of the sink to their target destinations.
    In this example, the pot cannot fit into the sink. 
    The second instance input has both dialogue and environment actions, and is evaluated at inference by whether the pot lands on the stove filled with water, and whether the potato is inside the pot and boiled.
    }
    \label{fig:edh_example}
\end{figure*}

\section{\teach Benchmarks}
\label{sec:benchmarks}

We introduce three benchmarks based on \teach\ sessions to train and evaluate the ability of embodied AI models to complete household tasks using natural language dialogue.
\textbf{Execution from Dialogue History} and \textbf{Trajectory from Dialogue} require modeling the \follower.
\textbf{Two-Agent Task Completion}, by contrast, requires modeling both the \commander\ and \follower\ agents to complete \teach\ tasks end-to-end.
For each benchmark, we define how we derive benchmark instances from \teach\ gameplay sessions, and by what metrics we evaluate model performance.

Each session has an initial state $S_i$, the sequence of actions $A = (a_1, a_2, \ldots )$ taken by the \commander\ and \follower\ including dialogue and environment actions, and the final state $S_f$.
We denote the subsequence of all dialogue actions as $A^D$, and of all navigation and interaction as $A^I$.
Following ALFRED, we create validation and test splits in both seen and unseen environments (Table~\ref{tab:splits})~\footnote{An earlier version of this dataset had a larger number of EDH instances. The current released split has filtered EDH instances so that only state changes that directly result in task progress are considered.}.
Seen splits contain sessions based in AI2-THOR rooms that were seen the training,
whereas unseen splits contain only sessions in rooms absent from the training set.

\begin{table}[t]
    \centering
    \begin{tabular}{llrr}
         Fold & Split & \# Sessions & \# EDH Instances \\
         \toprule
         Train & & 1482 (49\%) & 5475 (49\%) \\
         \midrule
         \multirow{2}{*}{Val} & Seen & 181 (\phantom{0}6\%) & 608 (\phantom{0}5\%) \\
          & Unseen & 612 (20\%) & 2157 (19\%) \\
         \midrule
         \multirow{2}{*}{Test} & Seen & 181 (\phantom{0}6\%) & 666 (\phantom{0}6\%) \\
          & Unseen & 589 (19\%) & 2270 (21\%) \\
         \bottomrule
    \end{tabular}
    \caption{Session and EDH instances in \teach\ data splits.}
    \label{tab:splits}
\end{table}

\subsection{Execution from Dialogue History (EDH)}

We segment \teach\ sessions into EDH instances.
We construct EDH instances $\left( S^{E}, A_H, A_R^I, F^{E} \right)$ where $S^{E}$ is the initial state of the EDH instance, $A_H$ is an action history, and the agent is tasked with predicting a sequence of actions that changes the environment state to $F^{E}$, using $A_R^I$ reference interaction actions taken in the session as supervision.
We constrain instances to have $|A_H^D|>0$ and at least one object interaction in $A_R^I$.
Each EDH instance is punctuated by a dialogue act starting a new instance or the session end.
We append a \action{Stop} action to each $A_R^I$.
An example is included in Figure~\ref{fig:edh_example}.
\begin{table*}[t]
    \centering
    \begin{tabular}{lrrrrrrrr}
         & \multicolumn{4}{c}{\textbf{EDH Validation}} & \multicolumn{4}{c}{\textbf{EDH Test}} \\
         & \multicolumn{2}{c}{\textit{Seen}} & \multicolumn{2}{c}{\textit{Unseen}} & \multicolumn{2}{c}{\textit{Seen}} & \multicolumn{2}{c}{\textit{Unseen}} \\
         Model & \multicolumn{1}{c}{SR [TLW]} & \multicolumn{1}{c}{GC [TLW]} & \multicolumn{1}{c}{SR [TLW]} & \multicolumn{1}{c}{GC [TLW]} & \multicolumn{1}{c}{SR [TLW]} & \multicolumn{1}{c}{GC [TLW]} & \multicolumn{1}{c}{SR [TLW]} & \multicolumn{1}{c}{GC [TLW]} \\
         \toprule
  Random & 0.82 [0.62] & 0.75 [0.43] & 1.34 [0.43] & 0.41 [0.07] & 0.6 [0.09] & 0.7 [0.27] & 1.89 [0.94] & 0.72 [0.16] \\
Lang & 0.99 [0.28] & 1.04 [0.29] & 2.36 [0.23] & 0.78 [0.29] & 0.75 [0.27] & 0.8 [0.31] & 2.03 [0.29] & 0.82 [0.14] \\
Vision & 5.1 [1.15] & 6.96 [1.76] & 3.89 [0.61] & 3.56 [0.73] & 5.86 [0.32] & 8.14 [1.09] & 4.23 [0.56] & 4.71 [0.66] \\
E.T. & 5.76 [0.90] & 7.99 [1.65] & 4.96 [0.54] & 4.71 [0.53] & 4.8 [1.07] & 8.08 [1.94] & 5.02 [0.91] & 5.57 [1.09] \\
+H & 7.4 [1.06] & 10.31 [2.02] & 4.31 [0.63] & 4.51 [0.72] & 6.01 [0.27] & 10.33 [1.58] & 5.68 [0.56] & 6.34 [0.75] \\
+A & 10.2 [0.71] & 15.71 [4.07] & 5.56 [0.51] & 5.2 [0.77] & 7.06 [0.53] & 9.57 [1.44] & 5.24 [0.67] & 6.1 [1.01] \\
+S & 8.55 [1.69] & 12.84 [3.41] & 7.83 [0.89] & 9.07 [1.69] & 4.2 [0.30] & 6.37 [1.80] & 3.88 [0.26] & 4.86 [0.84] \\
+H+A & 8.39 [0.82] & 14.92 [3.03] & 6.12 [0.88] & 6.43 [1.12] & 6.91 [0.37] & 10.34 [1.54] & 5.11 [0.67] & 5.84 [1.17] \\
+H+S & 9.38 [1.22] & 15.97 [3.55] & 5.7 [0.50] & 6.38 [0.78] & 6.31 [0.33] & 9.68 [1.71] & 5.29 [0.58] & 6.19 [0.93] \\       
         \bottomrule
         & \multicolumn{4}{c}{\textbf{TfD Validation}} & \multicolumn{4}{c}{\textbf{TfD Test}} \\
         \toprule
         Rand & 0.00 [0.00] & 0.00 [0.00] & 0.00 [0.00] & 0.00 [0.00] & 0.00 [0.00] & 0.00 [0.00] & 0.00 [0.00] & 0.00 [0.00] \\
         E.T. & \bf 1.02 [0.17] & \bf 1.42 [4.82] & \bf 0.48 [0.12] & \bf 0.35 [0.59] & \bf 0.51 [0.23] & \bf 1.60 [6.46] & \bf 0.17 [0.04] & \bf 0.67 [2.50] \\
         \bottomrule
    \end{tabular}
    \caption{E.T. outperforms random and unimodal baselines (\textbf{bold}).
    We ablate history loss (H), initializing with ALFRED (A), and initializing with ALFRED synthetic language (S).
    Metrics are success rate (SR) and goal condition success rate (GC).
    Trajectory length weighted metrics are included in [ brackets ]. 
    All values are percentages.
    For all metrics, higher is better.
    }
    \label{tab:edh_tfd_results}
\end{table*}

To evaluate inferred EDH action sequences, we compare the simulator state changes $\hat{E}$ at the end of inference with $F^{E}$ using similar evaluation criteria generalized from the ALFRED benchmark.
\begin{itemize}
    \item Success~$\{0,1\}$: 1 if all expected state changes $F^{E}$ are present in $\hat{E}$, else 0.
    We average over all trajectories.
    \item Goal-Condition Success (GC)~$(0, 1)$: The fraction of expected state changes in $F^{E}$ present in $\hat{E}$. 
    We average over all trajectories.\footnote{We follow ALFRED in using a macro-, rather than micro-average for Goal-Conditioned Success Rate.}
    \item Trajectory Weighted Metrics: For a reference trajectory $A_R^I$ and inferred action sequence $\hat{A}^I$, we calculate trajectory length weighted metric for metric value $m$ as
    $$
    TLW\mhyphen m = \frac{m * |A_R^I|}{max(|A_R^I|, |\hat{A}^I|)}.
    $$
\end{itemize}

During inference, the learned \follower\ agent predicts actions until either it predicts the \action{Stop} action, hits a limit of 1000 steps, or hits a limit of 30 failed actions.

\subsection{Trajectory from Dialogue (TfD)}

A \follower\ agent model is tasked with inferring the whole sequence of \follower\ environmental actions taken during the session conditioned on the dialogue history.
A TfD instance is $\left( S_i, A_H^D, A_R^I, S_f \right)$, where $A_H^D$ is all dialogue actions taken by both agents, and $A_R^I$ is all non-dialogue actions taken by the \follower.
We append a \action{Stop} action to $A_R^I$.
The agent does not observe dialogue actions in context, 
however, we 
use this task to test long horizon action prediction with a block of instructions, analogous to ALFRED or TouchDown~\cite{chen:cvpr19}.
We calculate success and goal-conditioned success by comparing $\hat{E}$ against state changes between $S_i$ and $S_f$.

\subsection{Two-Agent Task Completion (TATC)}

To explore modeling both a \commander\ and \follower\ agent, the TATC benchmark gives as input only environment observations to both agents.
The \commander\ model must use the \action{Progress Check} action to receive task information, then synthesize that information piece by piece to the \follower\ agent via language generation. 
The \follower\ model can communicate back via language generation.
The TATC benchmark represents studying the ``whole'' set of challenges the \teach\ dataset provides.
We calculate success and goal-conditioned success by comparing $\hat{E}$ against state changes between $S_I$ and $S_f$.


\section{Experiments and Results}

We implement initial baseline models and establish the richness of \teach\ data and difficulty of resulting benchmarks.

\subsection{\follower\ Models for EDH and TfD}

We use a single model architecture to train and evaluate on the EDH and TfD benchmark tasks.

\paragraph{Model.}
We establish baseline performance for the EDH and TfD tasks using the Episodic Transformer (E.T.) model~\cite{pashevich:arxiv21}, designed for the ALFRED benchmark. 
The original E.T. model trains a transformer language encoder and uses a ResNet-50 backbone to encode visual observations.
Two multimodal transformer layers are used to fuse information from the language, image, and action embeddings, followed by a fully connected layer to predict the next action and target object category for interaction actions.
E.T. uses a MaskRCNN~\cite{maskrcnn} model pretrained on ALFRED images to predict a segmentation of the egocentric image for interactive actions, matching the predicted mask to the predicted object category. 
We convert the centroid of this mask to a  relative coordinate specified to the \teach API wrapper for AI2-THOR. 

We modify E.T. by learning a new action prediction head to match \teach\ \follower\ actions.
Given an EDH or TfD instance, we extract all dialogue utterances from the action history $A_H^D$ and concatenate these with a separator between utterances to form the language input. 
The remaining actions $A_H^I$ are fed in order as the past action input with associated image observations.
Consequently, our adapted E.T. does not have temporal alignment between dialogue actions and environment actions.


Following the mechanism used in the original E.T. paper, we provide image observations from both actions in the history $A_H^I$, and the reference actions $A_R^I$, and task the model to predict the entire sequence of actions.
The model parameters are optimized using cross entropy loss between the predicted action sequence and the ground truth action sequence.
For EDH, we ablate a history loss (H) as cross entropy over the entire action sequence---actions in both $A_H^I$ and $A_R^I$, to compare against loss only against actions in $A_R^I$.
Note that in TfD, $|A_H^I|=0$.

We additionally experiment with initializing the model using weights trained on the ALFRED dataset. 
Note that since the language vocabulary and action space change, some layers need to be retrained.
For EDH, we experiment with initializing the model both with weights from the E.T. model trained only on base ALFRED annotations (A) and the model trained on ALFRED augmented with synthetic instructions (S) (from \citet{pashevich:arxiv21}).
We also perform unimodal ablations of the E.T. model to determine whether the model is simply memorizing sequences from the training data~\cite{thomason:arxiv18}.

At inference time, the agent uses dialogue history as language input, and the environment actions in $A_H^I$ as past action input along with their associated visual observations.
At each time step we execute the predicted action, with predicted object coordinate when applicable, in the simulator.
The predicted action and resulting image observation are added to agent's input for the next timestep.
The appendix details model hyperparameters.


\paragraph{Results.}
Table~\ref{tab:edh_tfd_results} summarizes our adapted E.T. model performance on the EDH and TfD benchmarks.

We observe that all E.T. model conditions in EDH are significantly better than \texttt{Random} and \texttt{Lang-Only} condition on all splits on SR and GC, according to a paired two-sided Welch $t$-test with Bonferroni corrections.
Compared to the \texttt{Vision-Only} baseline, the improvements of the E.T. models are statistically significant on unseen splits, but not on seen splits.
Qualitatively, we observe that the \texttt{Random} baseline only succeeds on very short EDH instances that only include one object manipulation involving a large target object, for example placing an object on a \object{countertop}. 
The same is true of most of the successful trajectories of the \texttt{Lang-Only} baseline. 
The success rate of the \texttt{Vision-Only} baseline suggests that the E.T.-based models are not getting much purchase with language signal.
Notably, E.T. performs well below its success rates on ALFRED, where it achieves 38.24\% on the ALFRED test-seen split and 8.57\% on the ALFRED test-unseen split.
Additionally, although there appears to be a small benefit from initializing the E.T. model with pretrained weights from ALFRED, these differences are not statistically significant.  
\teach\ language is more complex, involving multiple speakers, irrelevant phatic utterances, and dialogue anaphora.

E.T. model performance on TfD is poor but non-zero, unlike a \texttt{Random} baseline.
We do not perform additional ablations for TfD given the low initial performance.
Notably, in addition to the complexity of language, TfD instances have substantially longer average trajectory length ($\sim$130) than those in ALFRED ($\sim$50).  

\subsection{Rule-based Agents for TATC}

\begin{table}[t]
    \centering
    \begin{tabular}{l@{}rrr}
         Task& Success & \multicolumn{1}{c}{Rule Agent} & \multicolumn{1}{c}{Human} \\
         (Shrtnd) & Rate & \multicolumn{1}{c}{Actions/Session} & \multicolumn{1}{c}{Actions/Session} \\
         \toprule
         \task{Plant} & 26.70 & 230.26$\pm$\phantom{0}54.65 & 67.93$\pm$\phantom{0}40.70 \\
         \task{Coffee} & 54.55 & 120.24$\pm$\phantom{0}66.55 & 72.29$\pm$\phantom{0}50.85 \\
         \task{Clean} & 52.98 & 182.38$\pm$\phantom{0}79.84 & 96.92$\pm$\phantom{0}71.31 \\
         \task{All X Y} & 52.91 & 126.82$\pm$\phantom{0}64.75 & 103.53$\pm$\phantom{0}80.97 \\
         \task{Boil} & 0.00 & \na & 130.13$\pm$\phantom{0}94.80 \\
         \task{Toast} & 0.00 & \na & 136.11$\pm$\phantom{0}70.73 \\
         \task{N Slices} & 22.51 & 248.77$\pm$\phantom{0}98.57 & 146.23$\pm$113.96\\
         \task{X One Y} & 50.98 & 150.09$\pm$\phantom{0}97.12 & 147.80$\pm$104.45 \\
         \task{Cooked} & 1.67 & 424.25$\pm$135.57 & 189.26$\pm$\phantom{0}87.90 \\
         \task{Sndwch} & 0.00 & \na & 241.61$\pm$100.86 \\
         \task{Salad} & 1.55 & 351.20$\pm$\phantom{0}82.09 & 253.94$\pm$130.09 \\
         \task{Bfast} & 0.00 & \na & 359.90$\pm$162.33 \\
         \midrule
         \bf Overall & 24.40 & 161.54$\pm$\phantom{0}92.00 & 164.65$\pm$130.89 \\
         \bottomrule
    \end{tabular}
    \caption{Rule-based agent policies were expansive enough to solve some simple tasks about half the time, while being unable to solve most compositional tasks at all.
    Note that TATC performance is not directly comparable to EDH or TfD due to two-agent modeling in TATC.
    }
    \label{tab:tatc}
\end{table}

In benchmarks like ALFRED, a PDDL~\cite{pddl} planner can be used to determine what actions are necessary to solve relatively simple tasks.
In VLN, simple search algorithms yield the shortest paths to goals.
Consequently, some language-guided visual task models build a semantic representation of the environment, then learn a hierarchical policy to execute such planner-style goals~\cite{blukis:arxiv21}.

Inspired by such planning-based solutions, we attempted to write a pair of rule-based \commander\ and \follower\ agents to tackle the TATC benchmark.
In a loop, the rule-based \commander\ executes a \action{Progress Check} action, then forms a language utterance to the \follower\ consisting of navigation and object interaction actions needed to accomplish the next sub-goal in the response.
Each sub-goal needs to be identified by the language template used to describe it, then a hand-crafted policy must be created for the rule-based \commander\ to reference.
For example, for the \task{Put All X On Y} task, all sub-goals are of the form ``X needs to be on some Y'' for a particular instance of object \task{X}, and so a rule-based policy can be expressed as ``navigate to the X instance, pick up the X instance, navigate to Y, put X down on Y.''
\commander\ utterances are simplified to sequences of action names with a one-to-one mapping to \follower\ actions to execute, with interaction actions including $(x,y)$ screen click positions to select objects.
The rule-based agents perform \textit{no learning}.

Table~\ref{tab:tatc} summarizes the success rate of these rule-based agents across task types.
Note that for the tasks \task{Boil Potato}, \task{Make Plate of Toast}, \task{Make Sandwich}, and \task{Breakfast}, sub-goal policies were not successfully developed.
The rule-based agents represent about 150 hours of engineering work to hand-craft subgoal policies.
While success rates could certainly be increased by increasing subgoal policy coverage and handling simulation corner cases, it is clear that, unlike ALFRED and navigation-only tasks, a planner-based solution is not reasonable for \teach\ data and the TATC benchmark.
The appendix contains detailed implementation information about the rule-based agents.


\section{Conclusions and Future Work}

We introduce \teachfull\ (\teach), a dataset of over 3000 situated dialogues in which a human \commander\ and human \follower\ collaborate in natural language to complete household tasks in the AI2-THOR simulation environment.
\teach\ contains dialogue phenomena related to grounding dialogue in objects and actions in the environment, varying levels of instruction granularity, and interleaving of utterances between speakers in the absence of enforced turn taking. 
We also introduce a task definition language that is extensible to new tasks and even other simulators.
We propose three benchmarks based on \teach.
To study \follower\ models, we define the Execution from Dialogue History (EDH) and Trajectory from Dialogue (TfD) benchmarks, and evaluate an adapted Episodic Transformer~\cite{pashevich:arxiv21} as an initial baseline.
To study the potential of \commander\ and \follower\ models, we define the Two-Agent Task Completion benchmark, and explore the difficulty of defining rule-based agents from \teach\ data.

In future, we will apply other ALFRED modeling approaches~\cite{blukis:arxiv21,kim:eai21,zhang:arxiv21,suglia:arxiv21} to the EDH and TfD \follower\ model benchmarks.
However, \teach\ requires learning several different tasks, all of which are more complex than the simple tasks in ALFRED.
Models enabling few shot generalization to new tasks will be critical for \teach\ \follower\ agents.
For \commander\ models, a starting point would be to train a \textit{Speaker} model~\cite{speaker-follower} on \teach\ sessions.
We are excited to explore human-in-the-loop evaluation of \commander\ and \follower\ models developed for TATC.


\section{Acknowledgements}

We would like to thank Ron Rezac, Shui Hu, Lucy Hu, Hangjie Shi for their assistance with the data and code release, and Sijia Liu for assistance with data cleaning. We would also like to thank Nicole Chartier, Savanna Stiff, Ana Sanchez, Ben Kelk, Joel Sachar, Govind Thattai, Gaurav Sukhatme, Joel Chengottusseriyil, Tony Bissell, Qiaozi Gao, Kaixiang Lin, Karthik Gopalakrishnan, Alexandros Papangelis, Yang Liu, Mahdi Namazifar, Behnam Hedayatnia, Di Jin, Seokhwan Kim and Nikko Strom for feedback and suggestions over the course of the project.  

\bibliography{References}

\appendix

We provide additional statistics about \teach\ (\S\ref{sec:app_stats}), a summary of the data collection procedure for \teach\ sessions (\S\ref{sec:app_collection}), additional details about EDH and TfD benchmark experiments (\S\ref{sec:app_edh_tfd}), additional details about the rule-based agents we implemented for the TTAC benchmark (\S\ref{sec:app_tatc}), an explanation of the task definition language that guides the \teach\ (\S\ref{sec:app_tdl}), and representative examples and qualitative analysis of \teach\ sessions (\S\ref{sec:app_examples}).

\section{Additional \teach\ Statistics}
\label{sec:app_stats}

During data collection, we aimed to obtain at least 50 sessions per task in the unseen validation and test splits.
The exact number finally obtained varies due to the success rate of annotators in completing the tasks, and replicability issues related to non-determinism in the simulator. 
The final number of sessions per task per split is included in table \ref{tab:games_per-task_per_split}.

\begin{table}[!ht]
    \centering
    \tabcolsep 2.5pt
    \begin{tabular}{lccc}
  & \multicolumn{3}{c}{\bf \# Sessions} \\
  & \bf train & \bf val-seen & \bf val-unseen \\
  \toprule 
\texttt{Make Coffee} & 145 & 18 & 57 \\
\texttt{Water Plant} & 91 & 11 & 63 \\
\texttt{Make Plate Of Toast} & 90 & 11 & 52 \\
\texttt{Boil Potato} & 86 & 10 & 44 \\
\texttt{N Slices Of X In Y} & 164 & 20 & 56 \\
\texttt{N Cooked Slices} & 107 & 13 & 53 \\
\texttt{Prepare Salad} & 176 & 22 & 53 \\
\texttt{Prepare Sandwich} & 111 & 13 & 51 \\
\texttt{Clean All X} & 178 & 22 & 59 \\
\texttt{Put All X In One Y} & 167 & 20 & 50 \\
\texttt{Put All X On Y} & 184 & 23 & 50 \\
\texttt{Prepare Breakfast} & 162 & 20 & 53 \\
\bottomrule 
    \end{tabular}
    \caption{Number of sessions per task per split. Test split information is not currently included}
    \label{tab:games_per-task_per_split}
\end{table}

We created unseen splits using the same floorplans as the split used in ALFRED to enable easy sharing of models between \teach\ and ALFRED. 
In the future, we also plan to explore generating entirely new floorplans and layouts to expand the test scene distribution with controllable generation methods~\cite{luminous}.

\begin{figure}[!ht]
    \centering
    \includegraphics[width=0.45\textwidth]{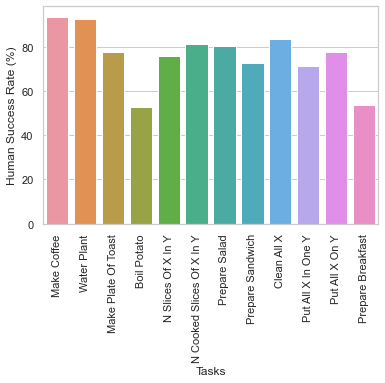}
    \caption{Human success rate for different tasks during data collection.
    Note that \teach\ benchmarks only contains successful dialogue sessions, so human performance here is more a measure of how complex tasks were for annotators to complete against both coordination and simulator quirks.}
    \label{fig:human_success}
\end{figure}

\begin{figure*}[!ht]
    \centering
    \includegraphics[width=0.95\textwidth]{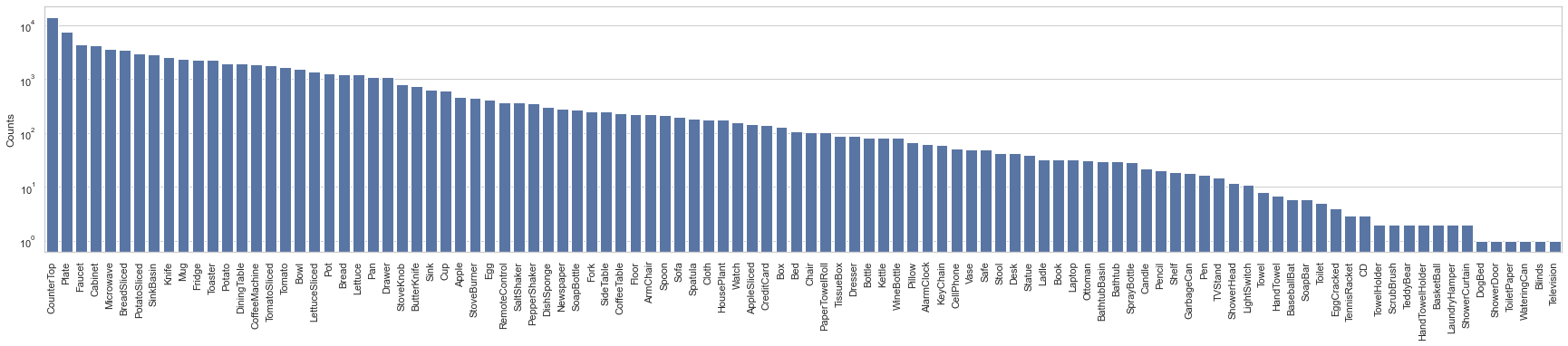}
    \caption{Object Distribution: Frequency with which objects are interacted with by the \follower\ across all sessions. Log scale.}
    \label{fig:object_dist}
\end{figure*}

The success rate of human annotators on different tasks when gathering data can be seen in Figure~\ref{fig:human_success}.
We find that success rates are much higher for simpler tasks such as \texttt{Make Coffee} and \texttt{Water Plant} compared to more difficult tasks. 
The lowest success rates were obtained with the \texttt{Prepare Breakfast} task which had the most steps, and consequently the maximum number of possible issues annotators could run into, and the \texttt{Boil Potato} task which required some additional reasoning from annotators in order to use a smaller container such as a \texttt{Cup} to fill a \texttt{Pot} or \texttt{Bowl} with water, in which the the \texttt{Potato} could then be boiled. 
Common causes of failure across tasks included difficulties in placement of objects (an artifact of AI2-THOR), objects being in initial positions where they are difficult to see and hence manipulate, connection problems, and timeouts due to one annotator becoming unresponsive.
Note that only sessions in which humans were successful are included in \teach\ benchmarks.

\begin{figure*}[!ht]
    \centering
    \subfloat[All]{
    \includegraphics[width=0.95\textwidth]{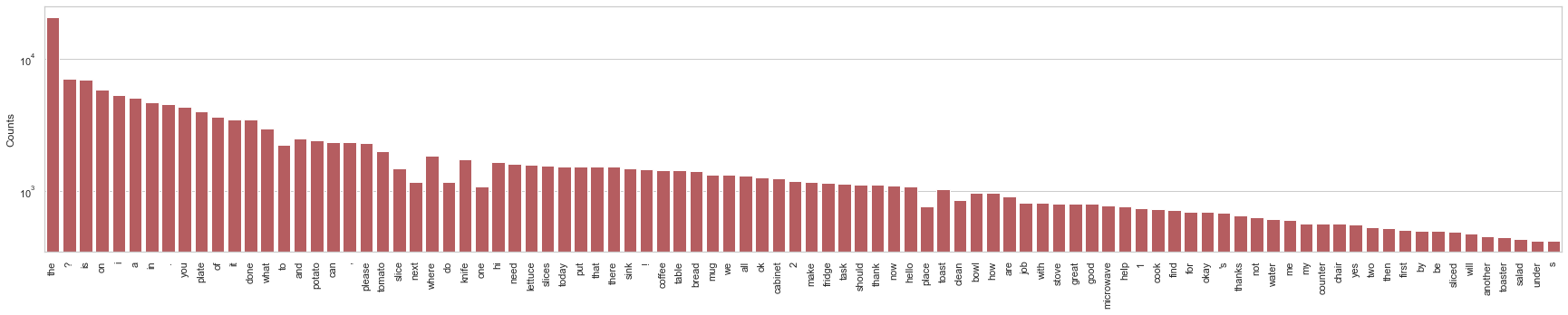}
    }
    
    \subfloat[Verbs]{
    \includegraphics[width=0.95\textwidth]{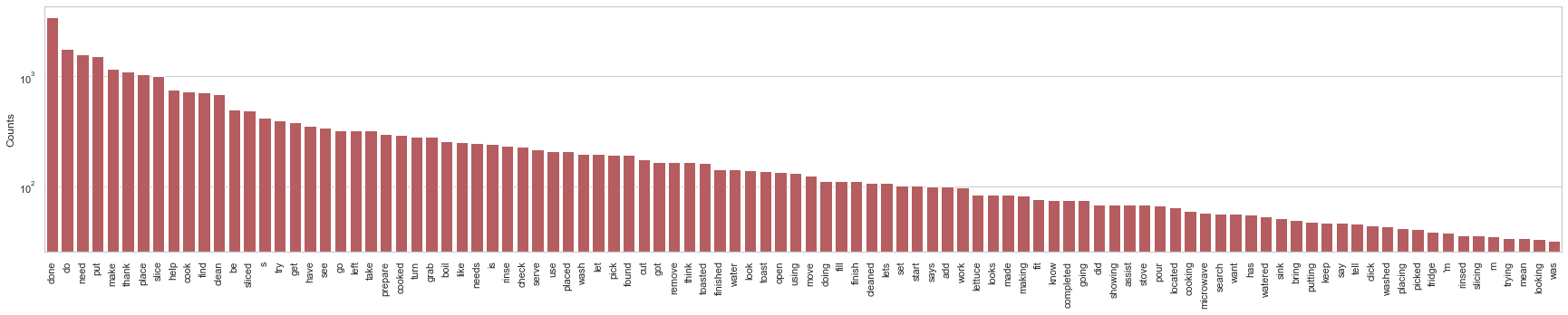}
    }
    
    \subfloat[Nouns]{
    \includegraphics[width=0.95\textwidth]{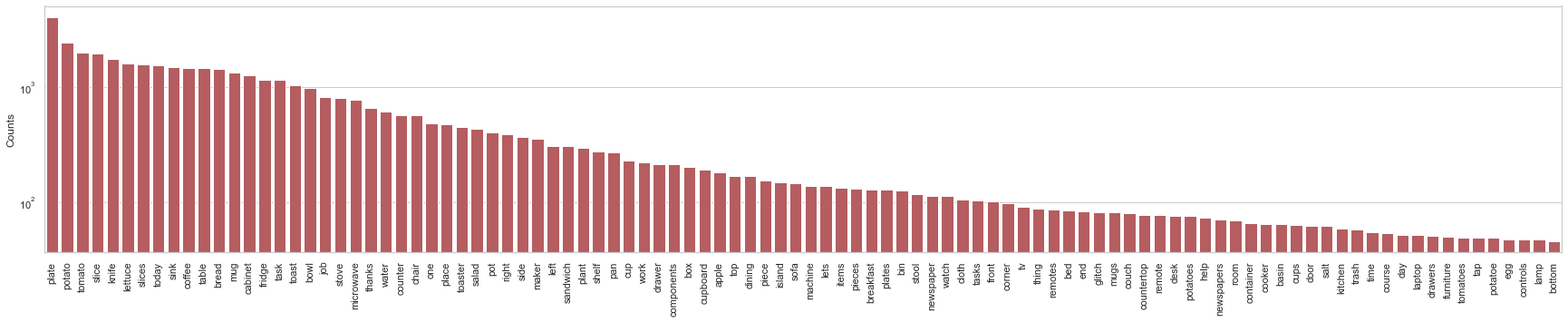}
    }
    
    \subfloat[Adjectives]{
    \includegraphics[width=0.95\textwidth]{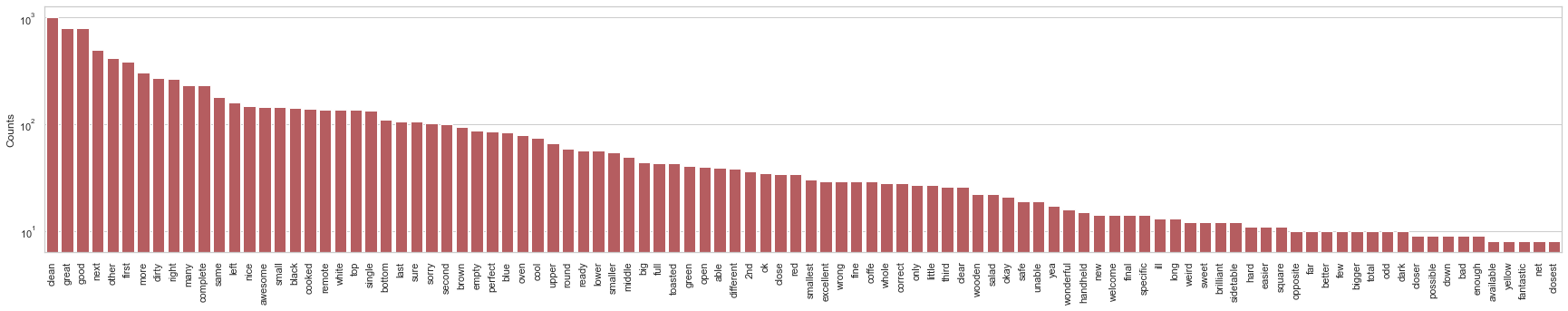}
    }
    \caption{Vocabulary Distributions: Frequency distributions of the 100 most common words, and 100 most common each of verbs, nouns and adjectives.
    Best viewed zoomed in.
    Log scale.}
    \label{fig:vocab}
\end{figure*}

We include vocabulary distributions of the 100 most common words in successful sessions in \teach, as well as the 100 most common each of verbs, nouns and adjectives in Figure~\ref{fig:vocab} (with POS tagging done using spaCy~\footnote{\url{https://pypi.org/project/spacy/}}).
We also include the distribution of the frequency with with objects of different types are interacted with by the \follower\ (Figure~\ref{fig:object_dist}). 
We can see that some objects get interacted with in many tasks, for example the counter is often used as a space to move things around, and faucets have to be interacted with frequently since many kitchen tasks involve the cleaning of utensils. 
Overall, since kitchen objects have more affordances, many of our tasks are set in the kitchen. 
The \texttt{Clean All X} task can be done in both the kitchen and the bathroom, but only the \texttt{Put All X On Y} and \texttt{Put All X In One Y} tasks can be done in the bedroom and living room.

Our task definition language is extensible.
We have defined \texttt{Prepare Study Desk}---analogous to \texttt{Prepare Breakfast} in scope and compositionality, for bedrooms and living rooms---together with some simpler tasks like \texttt{Turn On/Off All Lights} that better represent different room types.
We plan to incorporate these \textit{unseen tasks} into future iterations of the TATC benchmark.

\begin{figure}[!ht]
    \centering
    \includegraphics[width=\columnwidth]{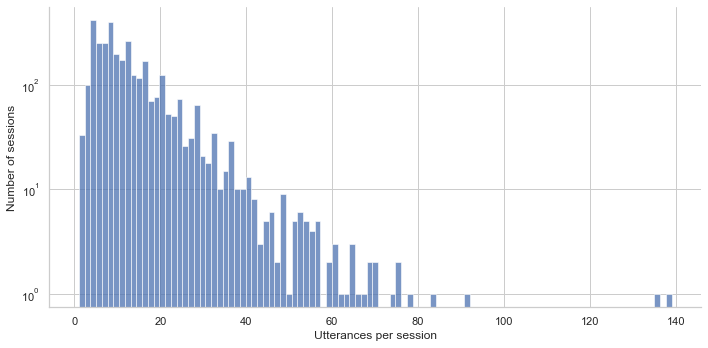}
    \caption{Distribution of dialogue lengths in terms of number of utterances per session. Log scale.}
    \label{fig:dialog_length_dist}
\end{figure}

\begin{figure}[!ht]
    \centering
    \includegraphics[width=\columnwidth]{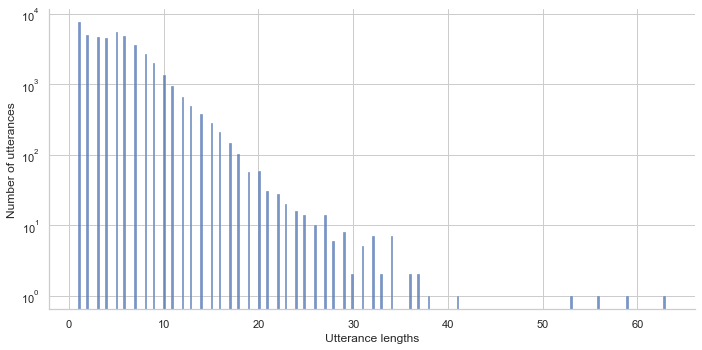}
    \caption{Distribution of utterance lengths in terms of number of tokens across sessions. Log scale.}
    \label{fig:utterance_length_dist}
\end{figure}

To analyze language in \teach, we include a distribution of the number of utterances per session in figure~\ref{fig:dialog_length_dist}.
We observe a significant number of games for a range of utterance lengths up to about 40 utterances per session, and the longest session has 139 utterances.
The distribution of utterance lengths can be seen in Figure~\ref{fig:utterance_length_dist}. 

\begin{figure}[!ht]
    \centering
    \includegraphics[width=\columnwidth]{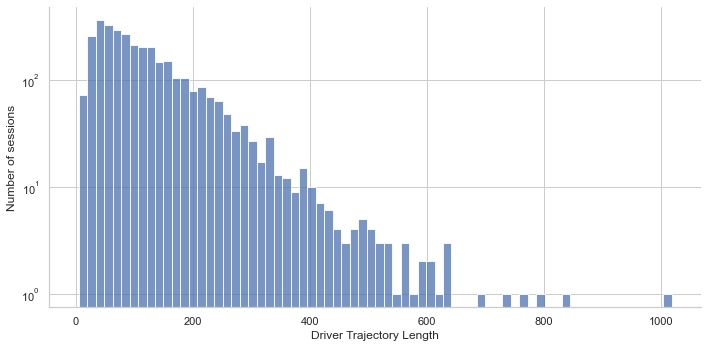}
    \caption{Distribution of \follower\ action trajectory lengths across sessions. Log scale.}
    \label{fig:game_traj_lens}
\end{figure}

We include a distribution of the number of environment actions taken by the \follower\ in in \ref{fig:game_traj_lens}. 
Our sessions are quite long, often involving several hundred actions per session.

\section{Data Cleaning}
In additional to original utterances entered by annotators in gameplay sessions, we also release cleaned versions of the utterances. Utterances were cleaned to remove spelling errors using SymSpell~\footnote{https://pypi.org/project/symspellpy/} followed by manual checking to avoid spurious changes, as well as expand contractions commonly used in chat (for example expanding ``nvm'' to ``never mind''). We also removed utterances that only referred to aspects of the annotation interface (for example the \commander\ mentioning that they did not understand what object was highlighted). Utterances that involved a mix of references to the interface and task relevant information were modified to retain task relevant information while removing references to the interface. All manual annotation was done by an expert annotator. While we did not used the cleaned utterances for modeling, we believe their availability will enable better generalization of trained models to utterances received in the form of speech, and to new utterances not collected via our interface.    

\section{Annotator Instructions}
\label{sec:app_collection}

Our annotator pool on Mechanical Turk was drawn using a private vendor service that provides high quality work in exchange for considerably higher pay than is normalized in the Mechanical Turk marketplace.

Annotators first completed a tutorial version of the task individually. 
In the tutorial, the annotator could see the tasks to be completed in the way the \commander\ does in the main annotation, but can also act in the environment as the \follower\ does.
They are then provided step by step instructions to control the \follower\ to complete two tasks - making coffee and cooking a slice of a potato. 
Only annotators who successfully completed the tasks in the tutorial were allowed to participate in the main collection.

Annotators were primed with the following description of the task:\\

\textit{\textbf{2-Player Game:}}

\textit{Now when you log into the game, you will be one of two players, the User/Commander or the Robot/Driver.} 

\textit{The User will have the progress check button but will not have the buttons to pick up / place objects or do other things with them.}

\textit{The Robot cannot see the progress check button but has the buttons to pick up / place objects or do other things with them.}

\textit{There is a chat box on the bottom right of the screen for the User and Robot to chat with each other. Both players have to work together to complete the task. }

\textit{\textbf{When you open the game, see if you have a Progress Check button on the top left.}} \\ 

\textit{\textbf{Robot}}:
\begin{itemize}
    \item \textit{If you don’t have a Progress Check button, you are the Robot.}
    \item \textit{You need to pretend you are a robot who is completing tasks in this house for your user.}
    \item \textit{Enter in the chat box ``What should I do today?'' so that your User knows you’re there.}
    \item \textit{Once the User tells you what to do, try to complete the task. You can use the chat to ask them questions, ask them to search for objects, or check whether steps have been completed. When the task is completed, your partner has to hit Finish to take both of you to the survey where you will rate your partner. When you submit the survey, you will get the code to enter in the HIT.  \\}
\end{itemize} 

\textit{\textbf{User}}:
\begin{itemize}
    \item \textit{When you open the game, if you have the Progress Check button, you are the User. }
    \item \textit{You need to pretend that the house in the game is your house and you are telling your Robot to complete tasks for you.}
    \item \textit{Remember that “the robot” is another worker like you pretending to be a robot to be respectful when talking to them.}
    \item \textit{To get started, click on the Progress Check button to see what tasks you have to do. Each HIT will have a slightly different task. Use the chat box to tell your Robot what task to do. }
    \item \textit{The Robot can ask you questions as they do the task. You can search for objects to help them and confirm whether they have successfully completed the task. }
    \item \textit{When the Progress Check button says that the task is done, hit Finish. That will take both you and your partner to the survey where you will rate each other. When you submit the survey, you will get the code to enter in the HIT. \\}
\end{itemize}

\textit{\textbf{Robot/ Driver Do’s and Don’ts}}

\textit{When you finish a step, use the chat box to tell the User and ask for the next step. E.g.: ``I toasted the bread. What should I do next?''}

\textit{When you need an object, ask the User to search for it first. If they cannot find it you will have to search for it yourself. Remember to open drawers and cabinets while searching.  For example: \\ \\
User: We need to make toast \\
Robot: Can you help me find the bread? \\
User: The bread is inside the fridge. \\
Robot can directly go to the fridge and get the bread. \\
Robot: I also need a knife. Do you know where that is? \\
User: I don’t know. Can you search for it? \\
Robot should search for knife. \\
} 

\textit{\textbf{User/ Commander Do’s and Don’ts}}

\textit{Use the search function in the Progress Checker to help the Driver find objects. If you don’t understand what it says you can say you don’t know. For example: \\ \\
User: We need to make toast \\
Robot: Can you help me find the bread? \\
User: The bread is inside the fridge. \\
Robot: I also need a knife. Do you know where that is? \\
User: I don’t know. Can you search for it? \\}

\textit{If a task has many steps. Don’t tell all of them as once. Wait for your Robot to finish a step before giving them the next step. \\
Good example: \\ \\
User: Today we need to put all the forks in the sink.  You can start with the one inside the microwave. \\
Robot: I got the fork from the microwave. Heading to the sink now. \\
Robot: I placed that fork in the sink. Are there more? \\
User: There is another fork on a plate to the left of the stove. \\
Robot: Found it. I will take it to the sink now. \\
Robot: That’s in the sink. What should I do next? \\
User: I think we’re done.  \\}

\textit{Bad example (\textbf{don't do this}): \\
User: Today we need to put all the forks in the sink. There is one inside the microwave and another on a plate to the left of the stove. \\}

\textit{Try to help the Robot solve problems. For example: \\}

\textit{User: Today we need to put all the forks in the sink. You can start with the one inside the microwave. \\
Robot: I found the fork but I am not able to place it in the sink. \\
User: Is the water running? \\
Robot: Yes it is \\
User: Try turning it off first \\
Robot: Thanks I tried that but I am still not able to place the fork in the sink \\
User: What else is there in the sink? \\
Robot: There is a plate and a few cups \\
User: Try removing something from the sink first. \\
}

After reading priming instructions, workers gathered 2-player sessions where they were randomly assigned one of the roles of \commander\ or \follower.

\begin{figure}[!ht]
    \centering
    \includegraphics[width=\columnwidth]{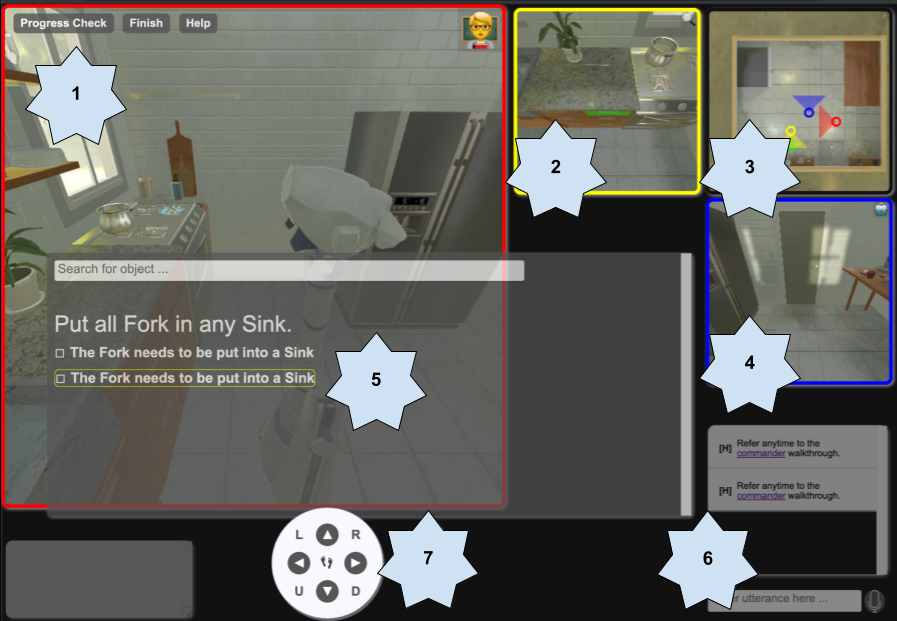}
    \caption{Interface seen by the crowdworker playing the \commander.
    Numbered stars are added for the purpose of explanation and correspond to the descriptions in this section.
    }
    \label{fig:commander_components}
\end{figure}

The interface seen by the \commander\ is shown in Figure~\ref{fig:commander_components}. The components of this interface are:
\begin{enumerate}
    \item Main panel: The \commander\ is allowed to move around in the environment. The Main Panel shows their egocentric view.
    \item Target object panel: If the \commander\ clicks on an instruction or searches for an object, a visual indication of the location of the object is shown here. For example in Figure~\ref{fig:commander_components}, one of the instruction steps is clicked and the target object panel is highlighting a drawer next to the sink. This indicates that a fork is present in that drawer which needs to be used in that instruction step. 
    \item Top down view: A top down view of the room the \follower\ is in. The blue circle is the \follower's current position, and the translucent blue triangle represents the view cone of the \follower. The red circle and cone correspond to the position of the \commander.
    \item \follower\ view panel: This shows the current egocentric view of the \follower .
    \item Progress Check Menu: This is shown when the Progress Check button has been clicked. It shows a list of steps to be completed.
    \item Chat window to send messages to the \follower.
    \item The navigation control panel used by the \commander\ to move around. The \commander\ can move through walls and objects but cannot interact with objects. \\
\end{enumerate}

\begin{figure}[!ht]
    \centering
    \includegraphics[width=\columnwidth]{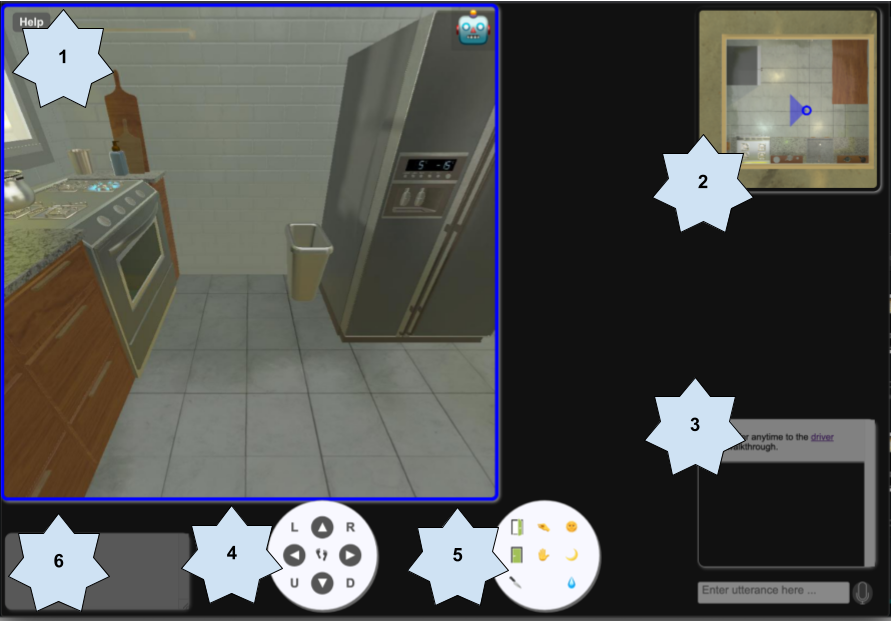}
    \caption{Interface seen by the crowdworker playing the \follower.
    Numbered stars are added for the purpose of explanation and correspond to the descriptions in this section.}
    \label{fig:driver_components}
\end{figure}

The interface used by the \follower\ is shown in Figure~\ref{fig:driver_components}. 
The components of this interface are:
\begin{enumerate}
    \item Main panel: This shows the egocentric view of the \follower.
    \item Top down view: A top down view of the room the \follower\ is in. The blue circle is the \follower's current position, and the translucent blue triangle represents the view cone of the \follower: the direction in the room the \follower\ is currently facing.
    \item Chat window to chat with the \commander.
    \item Navigation control panel to move the \follower\ and change its orientation.
    \item Object interaction control panel to interact with objects.
    \item Console log: This shows messages from the simulator to give annotators hints for what is going wrong when an action fails. For example, if the \follower\ tries to place an object but the placement fails, a simulator message might let the annotator know that the receptacle they're trying to place into is too full (e.g. they may need to clear out the sink before placing a new plate into it).
\end{enumerate}

A session is ended by the \commander\ clicking the ``Finish'' button adjacent to the ``Progress Check'' button on their interface, then confirming that they would like to end the game.
The ``Progress Check'' display changes to let the \commander\ know when the task is completed, prompting the \commander\ to end the game.
\teach\ contains only successful sessions where the task was completed.

\section{Additional EDH and TfD Experiment Details}
\label{sec:app_edh_tfd}

\begin{figure}[!ht]
    \centering
    \includegraphics[width=\columnwidth]{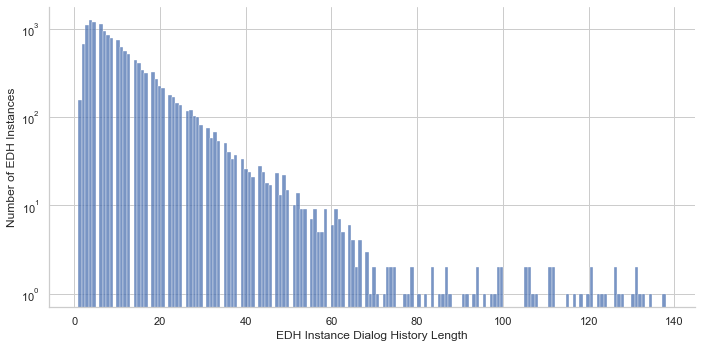}
    \caption{Distribution of dialog history length across EDH instances. Log scale.
    Note that EDH ``double counts'' many histories, skewing to a much longer, compounded average than full \teach\ sessions.}
    \label{fig:edh_dialog_history_lens}
\end{figure}

\begin{figure}[!ht]
    \centering
    \includegraphics[width=\columnwidth]{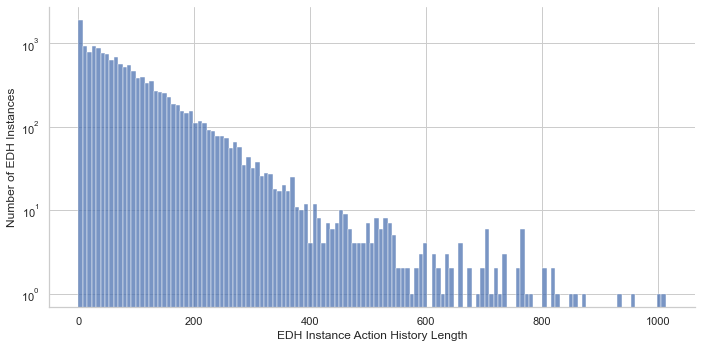}
    \caption{Distribution of action history lengths across EDH instances.}
    \label{fig:edh_action_history_lens}
\end{figure}

We include a distribution of EDH dialogue and action history lengths in Figures~\ref{fig:edh_dialog_history_lens} and \ref{fig:edh_action_history_lens}, respectively.
While the average action history length for EDH instances is $86.97$ actions, a significant number of EDH instances have an action history of over 200 actions. 

\begin{figure}[!ht]
    \centering
    \includegraphics[width=\columnwidth]{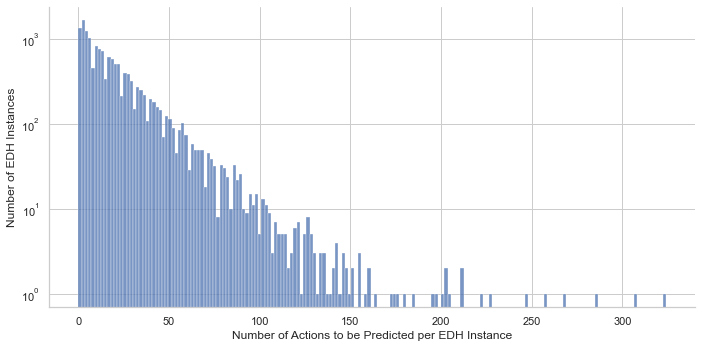}
    \caption{Distribution of number of actions to be predicted per instance across EDH instances.}
    \label{fig:pred_lens_per_edh}
\end{figure}

\begin{figure}[!ht]
    \centering
    \includegraphics[width=\columnwidth]{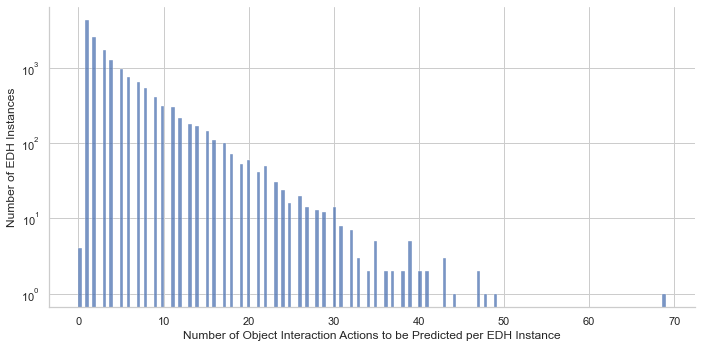}
    \caption{Distribution of number of object interaction actions to be predicted per instance across EDH instances.}
    \label{fig:obj_pred_lens_per_edh}
\end{figure}

We include the distribution of the number of all and object interaction actions to be predicted in Figures~\ref{fig:pred_lens_per_edh} and \ref{fig:obj_pred_lens_per_edh}, respectively.
While on average, a model for EDH needs to predict $19.76$ actions of which on average $4.74$ actions are object interaction actions, EDH instances may require as many as 324 actions to be predicted, with many instances requiring over 50 actions to be predicted. A significant number of EDH instances also require 10-20 object interactions to be predicted in order to successfully complete the instance.

\paragraph{Moeling EDH and TfD with E.T.}
For our human demonstrations, we showed image observations of size 900 x 900, and hence obtained images of the same size during replay for modeling. 
Images are resized to 224 x 244 for the ResNet-50 backbone of the main E.T. transformer and to 300 x 300 for the MaskRCNN model. 
We use the pretrained visual encoder based on Faster R-CNN and mask generator based on Mask R-CNN from E.T., where they are trained on 325K frames of expert demonstrations from the ALFRED train fold (which matches our train fold in terms of floorplans and objects visible). 
Additionally, as in E.T., we do not update the visual encoder or mask generator during model training for any of our tasks. 
The E.T. visual encoder average-pools ResNet features 4 times and adds a dropout of 0.3 to obtain feature maps of 512 x 7 x 7.
These are then fed into 2 convolutional layers of with 256 and 64 filters of size 1 x 1 respectively, and mapped using a fully connected layer to size 768.
Additionally, as in E.T., we use transformer encoders each with 2 blocks, 12 self attention heads, hidden size of 768, and dropout of 0.1. 
We also follow E.T in using the AdamW optimizer with 0.33 weight decay with a learning rate of $1e-4$ for the first 10 epochs and $1e-5$ for the last 10 epochs.
We trained all models for 20 epochs with a batch size of 3, and report results from the final epoch. 
We reused all hyperparameters except the batch size from the released E.T. model without further tuning, and used the largest batch size that could fit in a single GPU of a \texttt{p3.16x} EC2 instance. 
One change we made was that E.T. samples 30000 instances per epoch from the pool of training examples with replacement.
We replace sampling with rotation permutations of our training dataset per epoch, ensuring that every train example is seen exactly once in our dataset. 

E.T. uses 2 cross entropy losses: one over actions and one over object categories during object interaction actions.
We use an equal weight for these two losses. 
E.T. additionally uses auxiliary losses for overall and subgoal progress based on ALFRED but we do not use these as they are tailored for ALFRED and we do not have equivalent subgoal progress signals in \teach\ benchmarks.

Overall, our episode replay phase to generate image observations for training takes about 6 hours using 50 threads on a single \texttt{p3.16x} EC2 instance. 
The preprocessing phase of E.T. involving extracting image features and vectorizing language inputs takes about 7 hours using all GPUs of a single \texttt{p3.16x} EC2 instance. 
Training a model for 20 epochs takes about 5 hours using a single GPU of a single \texttt{p3.16x} EC2 instance. 
Evaluation of EDH instances takes about 6 hours using 2 GPUs of a single \texttt{p3.16x} EC2 instance for seen splits, and about 14 hours using 3 GPUs of a single \texttt{p3.16x} EC2 instance for unseen splits, but we did see a considerable amount of variation across runs.
TfD evaluation takes about 2 hours using 2 GPUs of a single  \texttt{p3.16x} EC2 instance for seen splits and 3 hours using 2 GPUs of a single \texttt{p3.16x} EC2 instance for unseen splits. 
We believe it should be possible to improve evaluation runtimes through optimizations to our wrapper over AI2-THOR. 

We include a breakdown of EDH success rates across tasks in table \ref{tab:per_task_success}. 
Note that the task referenced here is the task for the original gameplay session the EDH instance is created from.
Since our task definitions are hierarchical, it is possible for some EDH instances from different tasks to involve the same steps to be predicted.
For example, the \texttt{Make Coffee} task is a subtask of the \texttt{Prepare Breakfast} task, so it is possible for there to be EDH instances from sessions of the \texttt{Prepare Breakfast} task where the agent is only required to make coffee---the same as it would do in the \texttt{Make Coffee} task.

\begin{table*}[!ht]
\centering 
\small
\tabcolsep 3.5pt
    \centering
\begin{tabular}{lrrrrrrrrr}
& \multicolumn{9}{c}{\bf Per task EDH success rates on the valid-seen split} \\
Task & Rand & Lang & Vision & E.T. & +H & +A & +H+A & +S & +H+S\\ \toprule 
\texttt{Make Coffee} & 2.17 & 0.00 & 17.39 & 15.22 & 13.04 & 8.70 & 13.04 & 13.04 & 13.04\\
\texttt{Water Plant} & 0.00 & 0.00 & 0.00 & 10.53 & 10.53 & 15.79 & 10.53 & 21.05 & 5.26\\
\texttt{Make Plate Of Toast} & 0.00 & 2.38 & 0.00 & 7.14 & 4.76 & 4.76 & 2.38 & 7.14 & 4.76\\
\texttt{Boil Potato} & 0.00 & 0.00 & 0.00 & 17.39 & 8.70 & 13.04 & 4.35 & 13.04 & 13.04\\
\texttt{N Slices Of X In Y} & 2.08 & 1.04 & 7.29 & 14.58 & 8.33 & 10.42 & 8.33 & 12.50 & 8.33\\
\texttt{N Cooked Slices Of X In Y} & 1.15 & 4.60 & 4.60 & 8.05 & 4.60 & 6.90 & 5.75 & 10.34 & 8.05\\
\texttt{Clean All X} & 0.00 & 2.00 & 8.00 & 10.00 & 10.00 & 4.00 & 12.00 & 6.00 & 6.00\\
\texttt{Put All X In One Y} & 0.00 & 0.00 & 0.00 & 12.82 & 5.13 & 2.56 & 7.69 & 7.69 & 2.56\\
\texttt{Put All X On Y} & 0.00 & 0.00 & 6.25 & 4.17 & 10.42 & 8.33 & 8.33 & 10.42 & 6.25\\
\texttt{Prepare Salad} & 0.66 & 0.66 & 4.64 & 3.97 & 4.64 & 6.62 & 3.97 & 5.96 & 6.62\\
\texttt{Prepare Sandwich} & 1.02 & 1.02 & 4.08 & 6.12 & 5.10 & 5.10 & 9.18 & 6.12 & 8.16\\
\texttt{Prepare Breakfast} & 0.00 & 0.56 & 4.52 & 4.52 & 2.82 & 4.52 & 4.52 & 3.39 & 4.52\\
\bottomrule

& \multicolumn{9}{c}{\bf Per task EDH success rates on the valid-unseen split} \\
\toprule
\texttt{Make Coffee} & 2.00 & 0.00 & 4.00 & 8.00 & 9.00 & 7.00 & 8.00 & 7.00 & 10.00\\
\texttt{Water Plant} & 0.00 & 0.00 & 2.30 & 2.30 & 5.75 & 2.30 & 1.15 & 2.30 & 3.45\\
\texttt{Make Plate Of Toast} & 0.00 & 0.00 & 1.17 & 1.17 & 2.34 & 2.34 & 1.75 & 2.92 & 2.92\\
\texttt{Boil Potato} & 0.00 & 0.00 & 0.94 & 6.60 & 6.60 & 1.89 & 4.72 & 3.77 & 7.55\\
\texttt{N Slices Of X In Y} & 0.69 & 0.69 & 4.17 & 9.72 & 4.17 & 9.03 & 6.25 & 9.03 & 6.94\\
\texttt{N Cooked Slices Of X In Y} & 0.00 & 0.53 & 3.19 & 4.26 & 4.26 & 6.38 & 2.13 & 4.79 & 4.26\\
\texttt{Clean All X} & 0.91 & 0.00 & 4.55 & 3.64 & 9.09 & 3.64 & 6.36 & 4.55 & 2.73\\
\texttt{Put All X In One Y} & 1.02 & 0.00 & 2.04 & 6.12 & 1.02 & 4.08 & 3.06 & 3.06 & 3.06\\
\texttt{Put All X On Y} & 0.00 & 2.94 & 0.00 & 0.00 & 5.88 & 4.41 & 4.41 & 0.00 & 4.41\\
\texttt{Prepare Salad} & 0.75 & 0.38 & 1.13 & 6.04 & 2.26 & 4.53 & 1.51 & 3.40 & 3.02\\
\texttt{Prepare Sandwich} & 0.40 & 0.40 & 2.78 & 5.16 & 3.57 & 3.97 & 3.97 & 4.37 & 3.17\\
\texttt{Prepare Breakfast} & 0.27 & 0.82 & 4.37 & 5.19 & 6.28 & 3.28 & 5.19 & 6.56 & 4.37\\
\bottomrule

& \multicolumn{9}{c}{\bf Per task EDH success rates on the test-seen split} \\
\toprule 
\texttt{Make Coffee} & 0.00 & 0.00 & 4.17 & 16.67 & 14.58 & 10.42 & 10.42 & 12.50 & 12.50\\
\texttt{Water Plant} & 0.00 & 0.00 & 10.00 & 15.00 & 5.00 & 15.00 & 15.00 & 15.00 & 15.00\\
\texttt{Make Plate Of Toast} & 0.00 & 0.00 & 8.33 & 6.25 & 8.33 & 6.25 & 6.25 & 4.17 & 6.25\\
\texttt{Boil Potato} & 0.00 & 0.00 & 0.00 & 4.17 & 4.17 & 4.17 & 0.00 & 0.00 & 4.17\\
\texttt{N Slices Of X In Y} & 1.43 & 1.43 & 4.29 & 4.29 & 7.14 & 5.71 & 8.57 & 5.71 & 4.29\\
\texttt{N Cooked Slices Of X In Y} & 0.00 & 0.00 & 2.67 & 5.33 & 8.00 & 9.33 & 4.00 & 5.33 & 2.67\\
\texttt{Clean All X} & 4.05 & 0.00 & 6.76 & 8.11 & 8.11 & 6.76 & 16.22 & 6.76 & 13.51\\
\texttt{Put All X In One Y} & 2.00 & 6.00 & 0.00 & 4.00 & 6.00 & 4.00 & 8.00 & 4.00 & 6.00\\
\texttt{Put All X On Y} & 0.00 & 0.00 & 2.13 & 2.13 & 4.26 & 6.38 & 4.26 & 0.00 & 6.38\\
\texttt{Prepare Salad} & 1.38 & 0.00 & 4.83 & 6.90 & 4.83 & 6.90 & 4.14 & 6.90 & 6.21\\
\texttt{Prepare Sandwich} & 0.00 & 0.00 & 1.47 & 0.00 & 2.94 & 2.94 & 2.94 & 1.47 & 2.94\\
\texttt{Prepare Breakfast} & 0.00 & 1.75 & 4.68 & 4.68 & 6.43 & 6.43 & 8.77 & 4.09 & 6.43\\
\bottomrule

& \multicolumn{9}{c}{\bf Per task EDH success rates on the test-unseen split} \\
\toprule 
\texttt{Make Coffee} & 0.69 & 2.08 & 1.39 & 4.86 & 9.72 & 5.56 & 9.72 & 5.56 & 10.42\\
\texttt{Water Plant} & 0.00 & 0.00 & 0.00 & 0.00 & 0.00 & 0.00 & 0.00 & 0.00 & 0.00\\
\texttt{Make Plate Of Toast} & 0.37 & 0.75 & 4.10 & 5.22 & 3.36 & 4.10 & 5.97 & 4.10 & 5.22\\
\texttt{Boil Potato} & 0.53 & 0.53 & 4.28 & 3.21 & 2.67 & 6.95 & 3.74 & 6.42 & 3.21\\
\texttt{N Slices Of X In Y} & 0.97 & 0.49 & 3.88 & 4.85 & 6.80 & 3.88 & 4.85 & 6.31 & 5.34\\
\texttt{N Cooked Slices Of X In Y} & 1.87 & 1.49 & 1.87 & 7.84 & 3.73 & 6.34 & 4.85 & 8.58 & 4.48\\
\texttt{Clean All X} & 0.51 & 1.54 & 3.08 & 10.26 & 5.64 & 9.74 & 5.13 & 8.21 & 7.18\\
\texttt{Put All X In One Y} & 0.00 & 0.00 & 3.00 & 3.00 & 5.00 & 5.00 & 2.00 & 5.00 & 2.00\\
\texttt{Put All X On Y} & 0.00 & 0.00 & 0.00 & 5.22 & 5.22 & 5.22 & 5.22 & 2.24 & 2.24\\
\texttt{Prepare Salad} & 1.06 & 1.06 & 4.22 & 7.12 & 4.22 & 7.12 & 2.37 & 9.23 & 4.75\\
\texttt{Prepare Sandwich} & 0.00 & 1.01 & 3.36 & 6.04 & 4.36 & 4.36 & 3.02 & 5.03 & 2.35\\
\texttt{Prepare Breakfast} & 0.87 & 0.22 & 2.17 & 4.99 & 4.99 & 4.99 & 5.21 & 4.77 & 3.69\\
\bottomrule

\end{tabular}
\caption{Success rates on EDH benchmark divided by instance source task.
All values are percentages.}
\label{tab:per_task_success}
\end{table*}

\section{Rule-Based Agents for TATC}
\label{sec:app_tatc}

As mentioned in section 5.2, we engineered rule-based \commander\ and \follower\ agents as an attempt to solve the Two-Agent Task Completion benchmark. 

\textbf{Rule-based \follower} maintains a queue of actions that it needs to execute. Whenever this queue gets empty, it utters \emph{``What should I do next?"}.
\commander\ in the next turn detects this utterance and executes a \texttt{Progress Check} action to generate a templated language instruction. 
The templated instruction consists of space-separated low-level actions like \emph{``Forward Forward TurnRight LookUp Pickup Mug at 0.57 0.25"}. 
This instruction can be split up by the \follower\ into action tokens and each token has a one-to-one correspondence to an action in the action space of \follower.
For interaction action like \emph{``Pickup Mug at 0.57 0.25"}, \emph{``Mug''} represents the object to be interacted with, and \emph{`0.57 0.25"} represents the normalized coordinates in the egocentric view of \follower\ accepted by the \teach\ API as a click position.

\textbf{Rule-based \commander} executes a \texttt{Progress Check} action whenever it is asked to provide supervision. Listing~\ref{progresscheck:plate_of_toast} shows the \texttt{Progress Check} output at the start of \task{Make Plate of Toast} task. 
\texttt{\textbf{problem\_keys}} in the \texttt{Progress Check} output contains info about all objectives that need to be completed to solve the task. 
\texttt{\textbf{property\_name}} of a problem key defines the type of the problem key to be solved. For the 11 tasks we consider, there are 7 problem keys that can be solved: \texttt{parentReceptacle}, \texttt{isDirty}, \texttt{isFilledWithLiquid}, \texttt{isFilledWithCoffee}, \texttt{isBoiled}, \texttt{isCooked} and \texttt{Slice}. \texttt{\textbf{DesiredValue}} denotes the state that the object should be in. 
For each of the problem keys, we can engineer a hand-crafted logic to solve it.
Consequently, the scope of what rule-based agents can accomplish is limited by the engineering hours needed to identify and hand-write solutions to these problem keys in a modular planning fashion.

For \texttt{parentReceptacle} problem key, an object needs to be placed on/inside another object. Algorithm~\ref{alg:parentReceptacle} shows the hand-crafted logic for \texttt{parentReceptacle} problem key. 
Whenever \emph{Commander} is asked for supervision, it checks the \texttt{\textbf{property\_name}} of problem key and if it is \texttt{parentReceptacle}, it will call the \texttt{parentReceptacle()} function. 
The \texttt{current\_step} in the function keeps track of the supervision that needs to be provided based on the progress for the current problem key. For \texttt{current\_step = navigation\_1}, it will call a function \texttt{Navigate(ObjectID)} which runs a shortest path planner from the current pose of the \emph{Follower} agent to get the low-level navigation instruction to reach \texttt{ObjectID}. 
Similar logic can be written for other problem keys. 

\begin{algorithm}[!ht]
\caption{Handcrafted logic for \texttt{parentReceptacle}}
\label{alg:parentReceptacle}
\textbf{Input}: step\\
\textbf{Output}: instruction, step \\ \\
{parentReceptacle(step=``navigation\_1"):}
\begin{algorithmic}[1] 
\STATE ObjectID = get\_object\_id()
\STATE ParentID = get\_parent\_id() \% None if ObjectID not present inside an object
\IF {step = ``navigation\_1"}
\STATE instruction = Navigate(ObjectID)
\STATE step = ``interaction\_1\_1"
\ELSIF {step = ``interaction\_1\_1"}
\IF {ObjectID inside ParentObject}
\STATE instruction = ``ToggleOff ParentObject"
\STATE instruction += ``Open ParentObject"
\STATE step = ``interaction\_1\_2"
\ELSE
\STATE instruction = ``Pickup ObjectID"
\STATE step = ``step\_completed"
\ENDIF
\ELSIF {step = ``interaction\_1\_2"}
\STATE instruction = ``Pickup ObjectID"
\STATE step = ``interaction\_1\_3"
\ELSIF {step = ``interaction\_1\_3"}
\STATE instruction = ``Close ParentObject"
\STATE step = ``step\_completed"
\ENDIF
\STATE \textbf{return} instruction, step
\end{algorithmic}
\end{algorithm}

\begin{listing*}[!ht]
\begin{small}
\begin{minted}[frame=none,
               framesep=3mm,
               breaklines=true,
               xleftmargin=21pt,
               tabsize=1,
               tabsize=1,breaksymbolleft=]{json}
{
    "task_desc": "Make a plate of toast.",
    "success": 0,
    "subgoals": [
        {
            "representative_obj_id": "Bread|-00.58| 00.27|-01.27",
            "step_successes" : [0],
            "success": 0,
            "description": "Make a slice of toast.",
            "steps": [
                {
                    "success": 0,
                    "objectId": "Bread|-00.58| 00.27|-01.27",
                    "objectType": "Bread",
                    "desc": "The bread needs to be sliced using a knife.", },
                {
                    "success": 0,
                    "objectId": "Bread|-00.58| 00.27|-01.27",
                    "objectType": "Bread",
                    "desc": "The bread needs to be toasted.", } ],
            "problem_keys": {
                "Bread|-00.58| 00.27|-01.27" : [
                    {
                        "objectType": "Bread",
                        "determiner": "a",
                        "property_name": "objectType",
                        "desired_property_value": "BreadSliced" },
                    {
                        "objectType": "Bread",
                        "determiner": "a",
                        "property_name": "isCooked",
                        "desired_property_value": 1 } ] } },
        {
            "representative_obj_id": "Plate|-01.18| 00.21|-01.27",
            "step_successes": [1, 0],
            "success": 0,
            "description": "Clean a Plate.",
            "steps": [{
                    "success": 0,
                    "objectId": "Plate|-01.18| 00.21|-01.27",
                    "objectType": "Plate",
                    "desc": "The Plate is dirty. Rinse with water." }],
            "problem_keys": {
                "Plate|-01.18| 00.21|-01.27": [{
                        "objectType": "Plate",
                        "determiner": "a",
                        "property_name": "isDirty",
                        "desired_property_value": 0 }] } } ] }
\end{minted}
\end{small}
\caption{Sample \texttt{Progress Check} response for \task{Make a Plate of Toast}}
\label{progresscheck:plate_of_toast}
\end{listing*}

\section{Task Definition Language}
\label{sec:app_tdl}
We define a Task Definition Language to define household tasks in terms of object properties that need to be satisfied in the environment for the task to be considered successful. 
This Task Definition Language is based on a PDDL-like syntax~\cite{pddl}. 
A sample task can be seen in Listing~\ref{task:plate_of_toast}, which defines the \textsc{Make a Plate of Toast} task in \teach. 

\begin{listing}[!ht]
\begin{minted}[frame=none,
               framesep=3mm,
               breaklines=true,
               xleftmargin=21pt,
               tabsize=4,
               tabsize=2,breaksymbolleft=]{json}
{
  "task_id": 106,  
  "task_name": "Plate Of Toast",  
  "task_nparams": 0,  
  "task_anchor_object": "plate",  
  "desc": "Make a plate of toast.",  
  "components": {
    "toast": {
      "determiner": "a",      
      "task_name": "Toast",      
      "task_params": []
    },    
    "plate": {
      "determiner": "a",      
      "task_name": "Clean X",      
      "task_params": ["Plate"]
    }
  },  
  "relations": [
    {
      "property": "parentReceptacles",      
      "tail_entity_list": ["plate"],      
      "tail_determiner_list": ["the"],      
      "head_entity_list": ["toast"],      
      "head_determiner_list": ["a"],      
      "failure_desc": "The toast needs to be on a clean plate."    
    }
  ]
}
\end{minted}
\caption{Sample task: Make a Plate of Toast}
\label{task:plate_of_toast}
\end{listing}

A task is specified in terms of \texttt{\textbf{components}} and \texttt{\textbf{relations}}. 
A \texttt{\textbf{component}} is specified using a set of conditions to be satisfied for the task to be considered complete. 
As seen in the above example, a \texttt{\textbf{component}} can be specified by referencing another task.
In \textsc{Make a Plate of Toast}, the \texttt{\textbf{component}} \texttt{toast} is described by referencing another task, \texttt{Toast} (Listing~\ref{task:toast}), and the \texttt{\textbf{component}} \texttt{plate} is described by referencing another task, \texttt{Clean X} (Listing~\ref{task:clean_x}), with parameter value \texttt{Plate}.
In task definitions, \texttt{\textbf{relations}} are used to describe relationships between \texttt{\textbf{components}} that must be satisfied for the task to be considered complete. 
In \textsc{Make a Plate of Toast}, the relation specifies that one object satisfying the conditions of \texttt{\textbf{component}} \texttt{plate} must be the container of (captured by AI2-THOR property \texttt{parentReceptacles}) one object satisfying the conditions of \texttt{\textbf{component}} \texttt{toast}.

\begin{listing}[!ht]
\begin{minted}[frame=none,
               framesep=3mm,
               breaklines=true,
               xleftmargin=21pt,
               tabsize=4,
               tabsize=2,breaksymbolleft=]{json}
{
  "task_id": 101,  
  "task_name": "Toast",  
  "task_nparams": 0,  
  "task_anchor_object": "toast",  
  "desc": "Make a slice of toast.",  
  "components": {
    "toast": {
      "determiner": "a",      
      "primary_condition": "objectType",      
      "instance_shareable": false,      
      "conditions": {
        "objectType": "BreadSliced",        
        "isCooked": 1      },      
      "condition_failure_descs": {
        "objectType": "The bread needs to be sliced using a knife.",        
        "isCooked": "The bread needs to be toasted."      
      }
    }, 
    "knife": {
      "determiner": "a",
      "primary_condition": "objectType",
      "instance_shareable": true,
      "conditions": {
        "objectType": "Knife"
      },
      "condition_failure_descs": {
      }
    }
  },  
  "relations": []
}
\end{minted}
\caption{Sample task: Make Toast}
\label{task:toast}
\end{listing}

\begin{listing}[!ht]
\begin{minted}[frame=none,
               framesep=3mm,
               breaklines=true,
               xleftmargin=21pt,
               tabsize=4,
               tabsize=2,breaksymbolleft=]{json}
{
  "task_id": 103,  
  "task_name": "Clean X",  
  "task_nparams": 1,  
  "task_anchor_object": "#0",  
  "desc": "Clean a #0.",  
  "components": {
    "#0": {
      "determiner": "a",      
      "primary_condition": "objectClass",      
      "instance_shareable": false,      
      "conditions": {
        "objectClass": "#0",        
        "isDirty": 0      
      },      
      "condition_failure_descs": {
        "isDirty": "The #0 is dirty. Rinse with water."      
      }
    },    
    "sink": {
      "determiner": "a",      
      "primary_condition": "objectType",      
      "instance_shareable": true,      
      "conditions": {
        "objectType": "Sink",        
        "receptacle": 1      
      },      
      "condition_failure_descs": {
      }
    }
  },  
  "relations": []
}
\end{minted}
\caption{Sample task: Clean X}
\label{task:clean_x}
\end{listing}

The task \textsc{Clean X} is an example of a parameterized task.
The parameter \texttt{\#0} is set to \texttt{Plate} when this task is referenced as a part of the more complex task \textsc{Plate of Toast} (Listing~\ref{task:plate_of_toast}). 
In \textsc{Clean X}, the parameter is intended to be a custom object class (such as \texttt{Plate} or \texttt{Utensil}) which have been pre-defined. 
Parameters can also be used to specify in determiners or even free text to go into a natural language description, for example ``Put all Fork \textit{in} any Sink'' versus ``Put all cups \textit{on} any Table.''
When we check for task completion, parameters can be thought of as macros---we first do a text replacement of the parameter value wherever the parameter occurs in the task definition, and then the task definition is processed as if it does not have parameters. 
The use of parameters allows easy creation of different variants of a task with low manual effort, thus allowing us to create a more diverse dataset. 

More formally, a \textsc{Task} is defined by 
\begin{itemize}
    \item \textbf{\texttt{task\_id}} - A unique ID for the task
    \item \textbf{\texttt{task\_name}} - A unique name for the task used to reference it in other tasks
    \item \textbf{\texttt{task\_nparams}} - Number of parameters required by the task
    \item \textbf{\texttt{desc}} - Natural language prompt describing the task to provide to a \commander, (be it human or agent model.
    \item \textbf{\texttt{components}} - Dictionary specifying sets of conditions to be satisfied by objects of different types. This is used to specify both precondition objects required to complete the task such as knife if slicing is required or a sink if cleaning is required,  as well as objects that need to be converted to the correct state as part of the task such as \texttt{toast} in Listing~\ref{task:toast}. A component can also be described using another \texttt{Task}, such as \textsc{Clean X} being used to define the target receptacle on which toast should sit in \textsc{Make a Plate of Toast}. 
    \item \textbf{\texttt{relations}} - List of conditions that relate one set of objects to another. Currently the only relationship used in our task definitions is \texttt{parentReceptacles} which checks if one object is the container for other objects.
    However, pairwise operators could also be used to capture other spatial relations or time of completion of components. 
    \item \textbf{\texttt{task\_anchor\_object}} - This is either the key of a \textbf{\texttt{component}} or \texttt{null}. This is used to identify the specific object in the simulation environment whose properties would be checked when a \textbf{\texttt{component}} specified by this \textbf{\texttt{Task}} is used in a \textbf{\texttt{relation}}. For example, the \textsc{Make a Plate of Toast Task} (Listing~\ref{task:plate_of_toast}) contains a \textbf{\texttt{relation}} that says that its \texttt{toast} \textbf{\texttt{component}} should be contained in its \texttt{plate} \textbf{\texttt{component}}. Looking at the task definition for task \texttt{Toast} (Listing~\ref{task:toast}), we find that its \textbf{\texttt{task\_anchor\_object}} is its \textbf{\texttt{component}} \texttt{toast} (and not its \textbf{\texttt{component}} \texttt{knife}) which will resolve to an object of type \texttt{BreadSliced}. Looking at the task definition for \textsc{Clean X} (Listing~\ref{task:clean_x}), we find that its \textbf{\texttt{task\_anchor\_object}} is the \textbf{\texttt{component}} whose key is set to the value of parameter \texttt{\#0} which will be resolved to an object of type \texttt{\#0}. Since \textsc{Make a Plate of Toast} passes the parameter value \texttt{Plate} to \textsc{Clean X}, the \texttt{plate} \textbf{\texttt{component}} in \textsc{Make a Plate of Toast} will resolve to an object of type \texttt{Plate}. Thus overall the relation should check for an object of type \texttt{BreadSliced} (which also satisfies other conditions specified in \textbf{\texttt{component}} \texttt{toast}) to be placed on the object of type \texttt{Plate} (which also satisfies other conditions specified in \textbf{\texttt{component}} \texttt{plate}). Note that if the \textbf{\texttt{task\_anchor\_object}} for a \texttt{Task} is \texttt{null}, it cannot be used in \textbf{\texttt{relations}} in other \texttt{Task}s (but can still be a \textbf{\texttt{component}}).
\end{itemize}

A \textbf{\texttt{component}} can be of one of two types:
\begin{itemize}
    \item Atomic \textbf{\texttt{component}} - A \textbf{\texttt{component}} that is specified in terms of base simulator properties, for example all \textbf{\texttt{component}}s of \texttt{Task}s \textsc{Toast} and \textsc{Clean X}.
    \item \texttt{Task} \textbf{\texttt{component}} - A \textbf{\texttt{component}} that is specified in terms of another \texttt{Task}, for example the components in \textsc{Makke a Plate of Toast}).
\end{itemize}

Atomic \textbf{\texttt{component}}s are specified using the following keys:
\begin{itemize}
    \item \textbf{\texttt{conditions}} - Set of \texttt{property : desired\_value} pairs for this \textbf{\texttt{component}} to be considered satisfied. For example the conditions for the \texttt{toast} \textbf{\texttt{component}} in \textsc{Toast} look for an object of type \texttt{BreadSliced} whose property \texttt{isCooked} has been set to 1.
    \item \textbf{\texttt{condition\_failure\_descs}} - For properties in conditions that correspond to changes that have to happen by the annotator taking an action, this specifies the description to be provided to the annotator if the property is currently not set to the desired value. For example, in \textbf{\texttt{component}} \texttt{toast} the value for \textbf{\texttt{condition\_failure\_descs}} specifies that if there is no sliced bread in the scene, we should send the message ``The bread needs to be sliced using a knife'' and if there is no toasted bread slice in the scene we should send the message ``The bread needs to be toasted''.
    \item \textbf{\texttt{determiner}} - This is used to specify how many object instances should satisfy this set of conditions. The possible values are \texttt{a}, \texttt{all} or a positive integer. For example in the \texttt{Task Toast}, the \texttt{toast} \textbf{\texttt{component}} has \textbf{\texttt{determiner}} \texttt{a} so we would say this component is satisfied if there is any slice of toasted bread in the scene. Instead if the \textbf{\texttt{determiner}} was 2, we would only say that the \textbf{\texttt{component}} is satisfied if there are at least 2 slices of toasted bread in the scene. If it was \texttt{all}, we would say that the \textbf{\texttt{component}} is satisfied if all slices of bread present in the scene are toasted.
    \item \textbf{\texttt{primary\_condition}} - This is used to find candidate objects that an annotator needs to modify to satisfy this \textbf{\texttt{component}}. It is usually an object type or class.
    \item \textbf{\texttt{instance\_shareable}} -  This is a parameter used to handle how numbers cascade across hierarchies. In the \textsc{Make a Plate of Toast}, the \textbf{\texttt{determiner}} for component \texttt{toast} is \texttt{a}. Suppose instead that this was 2. By default we would multiply the \textbf{\texttt{determiner}}s of all \textbf{\texttt{component}}s of \textsc{Toast} (except \texttt{all}) by 2 (treating \texttt{a} as 1). So to check if the \textsc{Toast-2} is satisfied we would check both if there are 2 slices of toast and if there are 2 knives. But the knife has only been specified as a precondition and we do not actually need 2 knives in \textsc{Toast-2}. This exception is captured by the property \textbf{\texttt{instance\_shareable}}. The knife \textbf{\texttt{component}} has \textbf{\texttt{instance\_shareable}}\texttt{ = true} so regardless of the \textbf{\texttt{determiner}} associated with \textsc{Toast},  we would only check for one knife, but the \textbf{\texttt{component}} \texttt{toast} has \textbf{\texttt{instance\_shareable}}\texttt{ = false} so we would require $n$ slices of toast in \textsc{Toast-2}.
\end{itemize}

\begin{listing}[!ht]
\begin{minted}[frame=none,
               framesep=3mm,
               breaklines=true,
               xleftmargin=21pt,
               tabsize=4,
               tabsize=2,breaksymbolleft=]{json}
{
  "task_id": 110,
  "task_name": "Put All X On Y",
  "task_nparams": 3,
  "task_anchor_object": null,
  "desc": "Put all #0 #1 any #2.",
  "components": {
    "#0": {
      "determiner": "all",
      "primary_condition": "objectClass",
      "instance_shareable": false,
      "conditions": {
          "objectClass": "#0"
      },
      "condition_failure_descs": {}
    },
    "#2": {
      "determiner": "a",
      "primary_condition": "objectClass",
      "instance_shareable": true,
      "conditions": {
          "objectClass": "#2",
          "receptacle": 1
      },
      "condition_failure_descs": {}
    }
  },
  "relations": [
    {
      "property": "parentReceptacles",
      "tail_entity_list": ["#2"],
      "tail_determiner_list": ["a"],
      "head_entity_list": ["#0"],
      "head_determiner_list": ["all"],
      "failure_desc": "The #0 needs to be put #1to a #2"
    }
  ]
}
\end{minted}
\caption{Sample task: Put All X On Y}
\label{task:put_all_x_on_y}
\end{listing}

\begin{listing}[!ht]
\begin{minted}[frame=none,
               framesep=3mm,
               breaklines=true,
               xleftmargin=21pt,
               tabsize=4,
               tabsize=2,breaksymbolleft=]{json}
{
  "task_id": 111,
  "task_name": "Put All X In One Y",
  "task_nparams": 3,
  "task_anchor_object": null,
  "desc": "Put all #0 #1 one #2.",
  "components": {
    "#0": {
      "determiner": "all",
      "primary_condition": "objectClass",
      "instance_shareable": false,
      "conditions": {
          "objectClass": "#0"
      },
      "condition_failure_descs": {}
    },
    "#2": {
      "determiner": "a",
      "primary_condition": "objectClass",
      "instance_shareable": true,
      "conditions": {
          "objectClass": "#2",
          "receptacle": 1
      },
      "condition_failure_descs": {}
    }
  },
  "relations": [
    {
      "property": "parentReceptacles",
      "tail_entity_list": ["#2"],
      "tail_determiner_list": ["the"],
      "head_entity_list": ["#0"],
      "head_determiner_list": ["all"],
      "failure_desc": "The #0 needs to be put #1to a single #2"
    }
  ]
}
\end{minted}
\caption{Sample task: Put All X In One Y}
\label{task:put_all_x_in_one_y}
\end{listing}

Task relations are specified by:
\begin{itemize}
    \item \textbf{\texttt{property}} - The property being checked (currently we only have support for \texttt{parentReceptacles}) 
    \item \textbf{\texttt{head\_entity\_list}}, \textbf{\texttt{tail\_entity\_list}} - While exactly which object is the head and which object is the tail is arbitrary and would be decided by implementation used to check a property, we assume that if we examine the property value of the head entities, the tail entities would be specified in them. Currently these are implemented as lists to handle the very specific case where we want to define that multiple objects need to be placed in a single container (e.g.: multiple sandwich components in a plate). The entities are specified using the \textbf{\texttt{component}} keys and we recursively check \textbf{\texttt{task\_anchor\_object}} of \texttt{Task} components to find the exact object to be used when checking the relation.
    \item \textbf{\texttt{head\_determiner\_list}} - A list of the same length as \textbf{\texttt{head\_entity\_list}} where each entry can take values \texttt{a}, \texttt{all}, or a number and specify how many objects matching conditions specified by the respective \textbf{\texttt{component}} are involved in this relation.
    \item \textbf{\texttt{tail\_determiner\_list}} - A list of the same length as \textbf{\texttt{tail\_entity\_list}} where each entry can take values \texttt{a} or \texttt{the}. To illustrate the difference, compare the \texttt{Task}s \textsc{Put All X On Y} (Listing~\ref{task:put_all_x_on_y}) and \textsc{Put All X In One Y} (Listing~\ref{task:put_all_x_in_one_y}). Suppose there are two objects of type \texttt{X} ($x_1$ and $x_2$) and two objects of type \texttt{Y} ($y_1$ and $y_2$) in the scene. If $x_1$ is placed in/on $y_1$ and $x_2$ is placed in/on $y_2$, this would satisfy the \textsc{Task Put All X On Y} (because each object of type \texttt{X} is on \texttt{a} object of type \texttt{Y}) but does not satisfy \textsc{Put All X In One Y} (because there is no single object of type \texttt{Y} such that $x_1$ and $x_2$ are in \texttt{the} object of type \textsc{Y})
    \item \textbf{\texttt{failure\_desc}} - The message to be shown to an annotator if some action needs to be taken to make this relation satisfied.
\end{itemize}

\textbf{Checking task completion}: The task definition specifies all conditions that need to be satisfied by objects in the scene for a \texttt{Task} to be considered satisfied.
To check if a \texttt{Task} is satisfied, first, for each \textbf{\texttt{component}}, we check if as many instances, specified by the  \textbf{\texttt{determiner}} of that  \textbf{\texttt{component}} satisfy the conditions specified in  \textbf{\texttt{conditions}} (or in the case of \texttt{Task}  \textbf{\texttt{component}}s, we recursively check that the \texttt{Task} specified as the  \textbf{\texttt{component}} is satisfied).  
Next, we take the objects satisfying the conditions of each  \textbf{\texttt{component}} and use them to check  \textbf{\texttt{relation}}s. 
If there exist objects within this subset that also satisfy all  \textbf{\texttt{relation}}s, the \texttt{Task} is considered satisfied.

\section{\teach\ Examples and Qualitative Analysis}
\label{sec:app_examples}

For several example figures below, we provide video session replays in the attached supplementary material.
To compress video size and length, we play 1 action, either an utterance or environment action, per second, rather than the ``real time'' playback.
Videos show the \commander\ and \follower\ egocentric view, as well as the object search camera for the \commander\ together with the segementation mask of the searched object.
Additionally, each video shows utterance data and progress check response data in JSON format.

\teach\ was collected using an annotator interface that allowed unconstrained chat between annotators.
Annotators need to communicate because the task information is only available to the \commander---during collection called the \textit{User}---but only the \follower---in collection called the \textit{Robot}---can actually take actions.
We provide some guidelines for annotators on how to conduct these conversations, detailed in \S\ref{sec:app_collection}.
Our guidelines encourage annotators to explicitly request for and mention only task-relevant information.
However, annotators can and do decide to provide annotations in different levels of detail and relevance.

\begin{figure*}[!ht]
    \centering
    \includegraphics[width=1\textwidth]{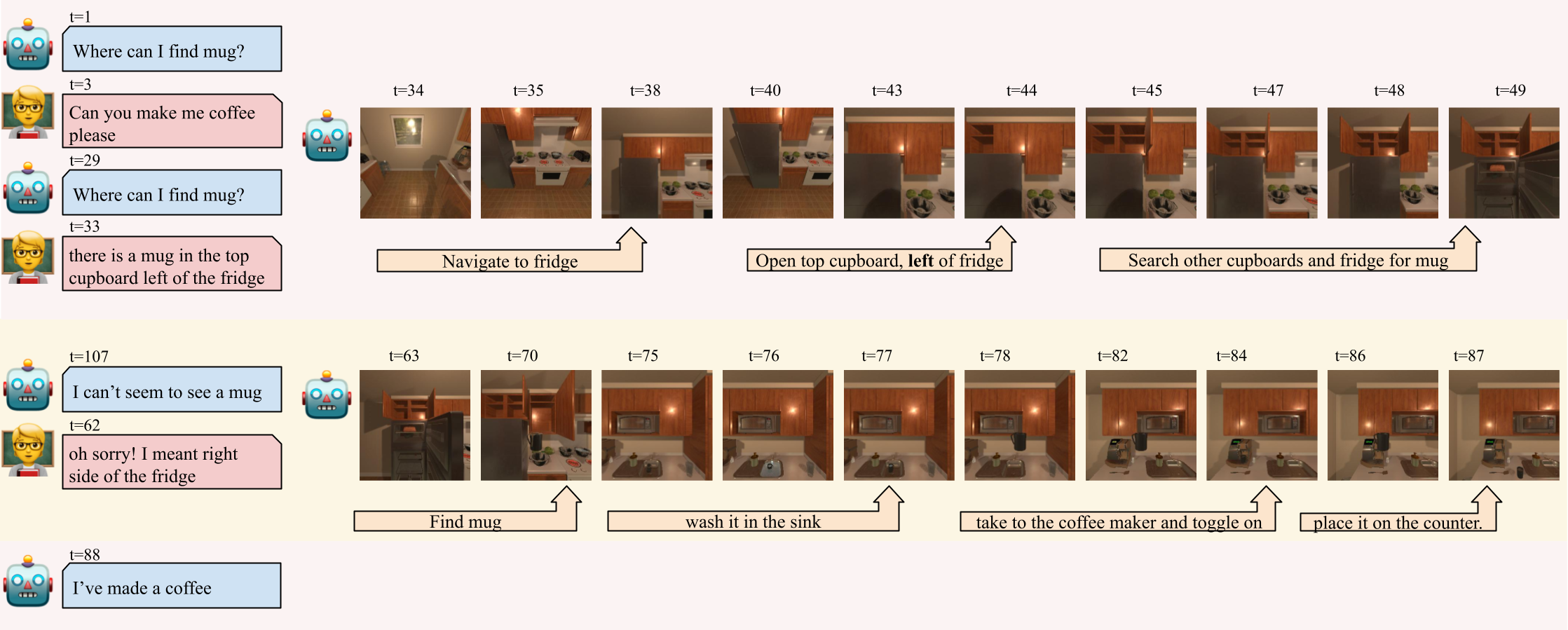}
    \caption{Sample session for the \texttt{Make Coffee} task where the \commander\ does not explain in much detail how the task is to be completed. 
    The session also includes an example where the \follower\ needs to ask for help because the \commander\ initially provided the wrong location for the mug.}
    \label{fig:sample_dialog_1}
\end{figure*}

\begin{figure*}[!ht]
    \centering
    \includegraphics[width=1\textwidth]{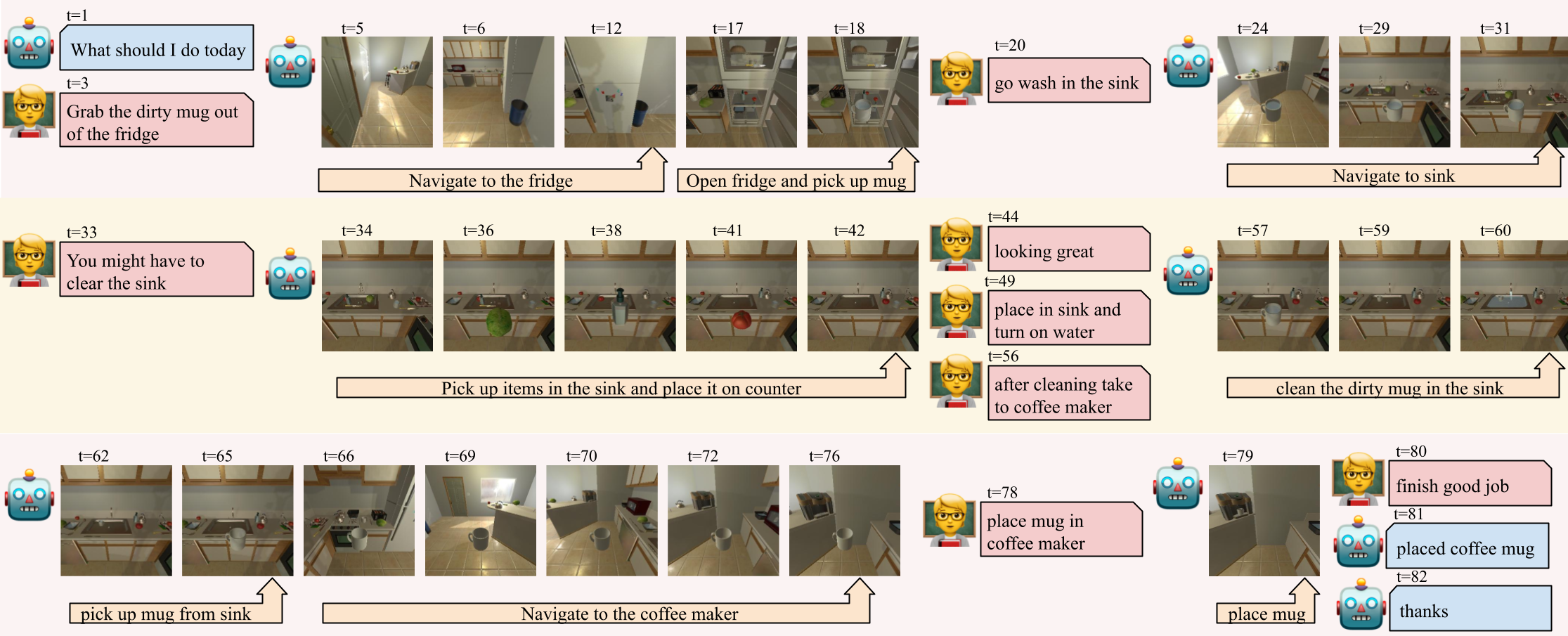}
    \caption{Another sample session for \texttt{Make Coffee}. 
    In this session, the \commander\ provides step by step instructions and feedback to the \follower\ despite the \follower\ not asking for the next instruction or help.}
    \label{fig:Coffee_example_2}
\end{figure*}

\begin{figure*}[!ht]
    \centering
    \includegraphics[width=1\textwidth]{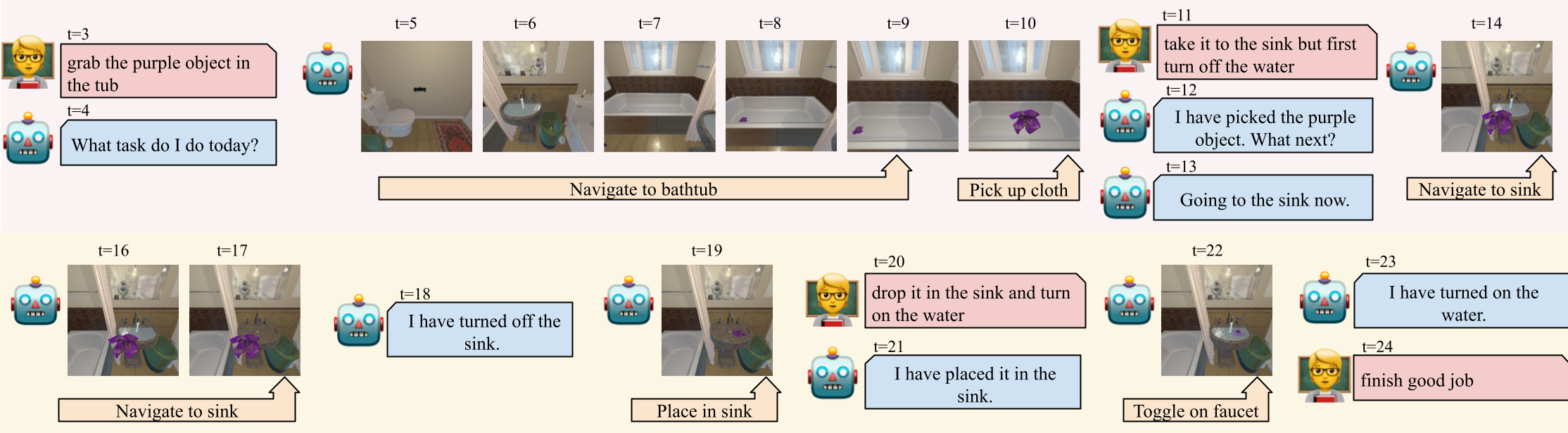}
    \caption{Sample session for the \texttt{Clean All X} task in a bathroom. 
    While the task could be solved more efficiently by simply turning on the faucet in the bathtub, the \commander\ and \follower\ instead choose to clean the cloth in the sink. 
    This session also demonstrates examples of how utterances can get out of order due to the absence of forced turn taking.
    }
    \label{fig:example_clean_all_x}
\end{figure*}

\begin{figure*}[!ht]
    \centering
    \includegraphics[width=1\textwidth]{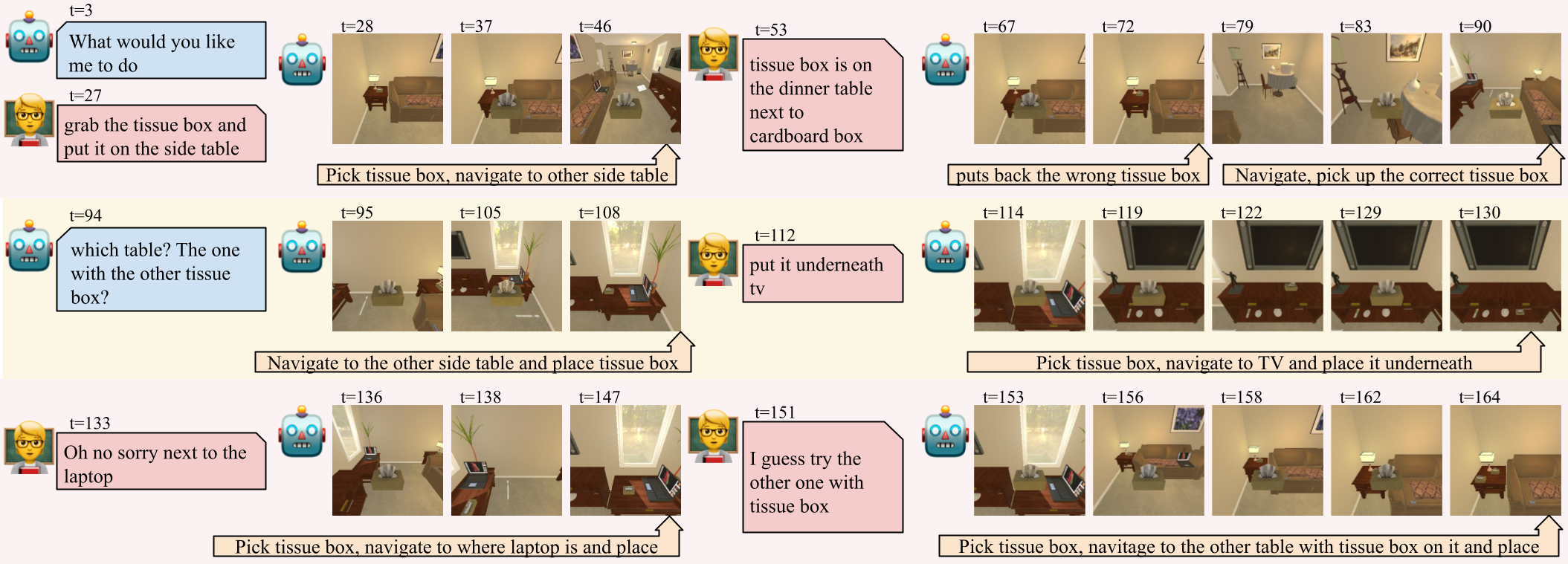}
    \caption{Sample session for the \texttt{Put All X In One Y} task. 
    In this session, \commander\ corrects the \follower\ to pick up the correct tissue box.
    Then \commander\ does not realize that all the tissue boxes need to be placed on the same side table, and hence initially gives the \follower\ incorrect instructions.}
    \label{fig:example_put_all_x_in_one_y}
\end{figure*}

\begin{figure*}[!ht]
    \centering
    \includegraphics[width=1\textwidth]{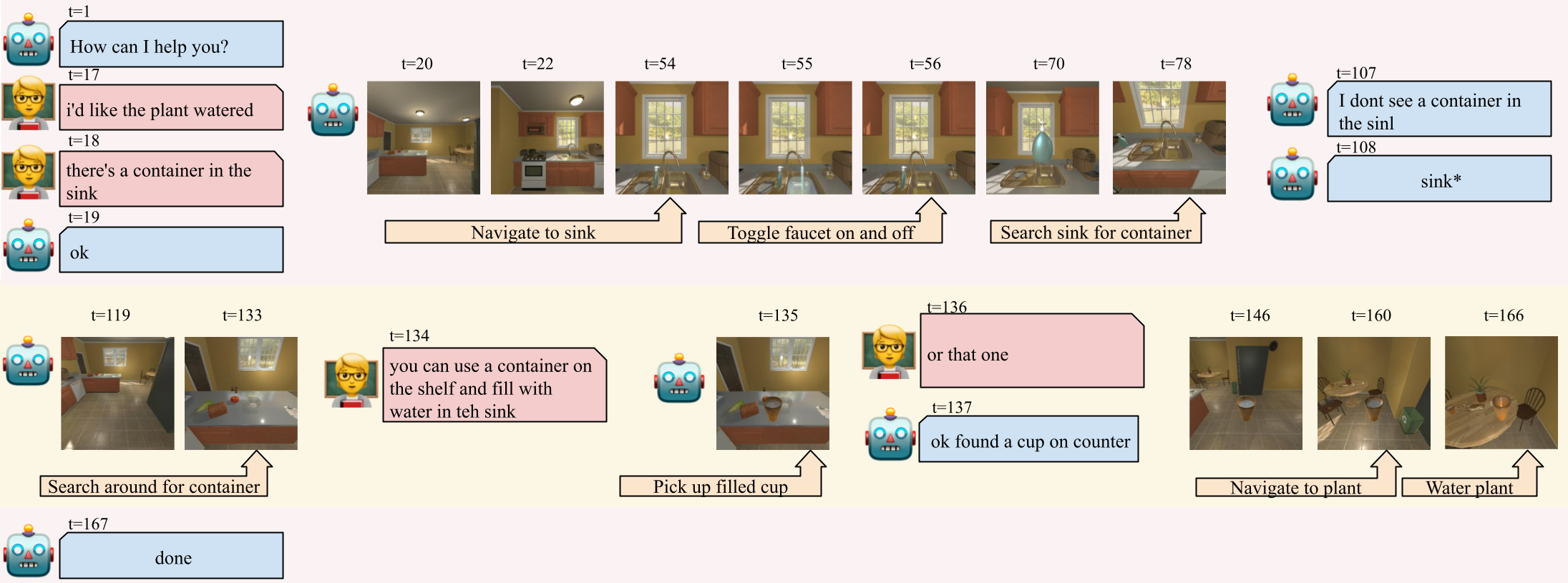}
    \caption{Sample session for the \texttt{Water Plant} task. 
    The \commander\ initially gives an incorrect instruction requiring the \follower\ to ask for help and search for a container. 
    The \follower\ finds another container before getting help from the \commander.}
    \label{fig:example_water_plant}
\end{figure*}

\begin{figure*}[!ht]
    \centering
    \includegraphics[width=1\textwidth]{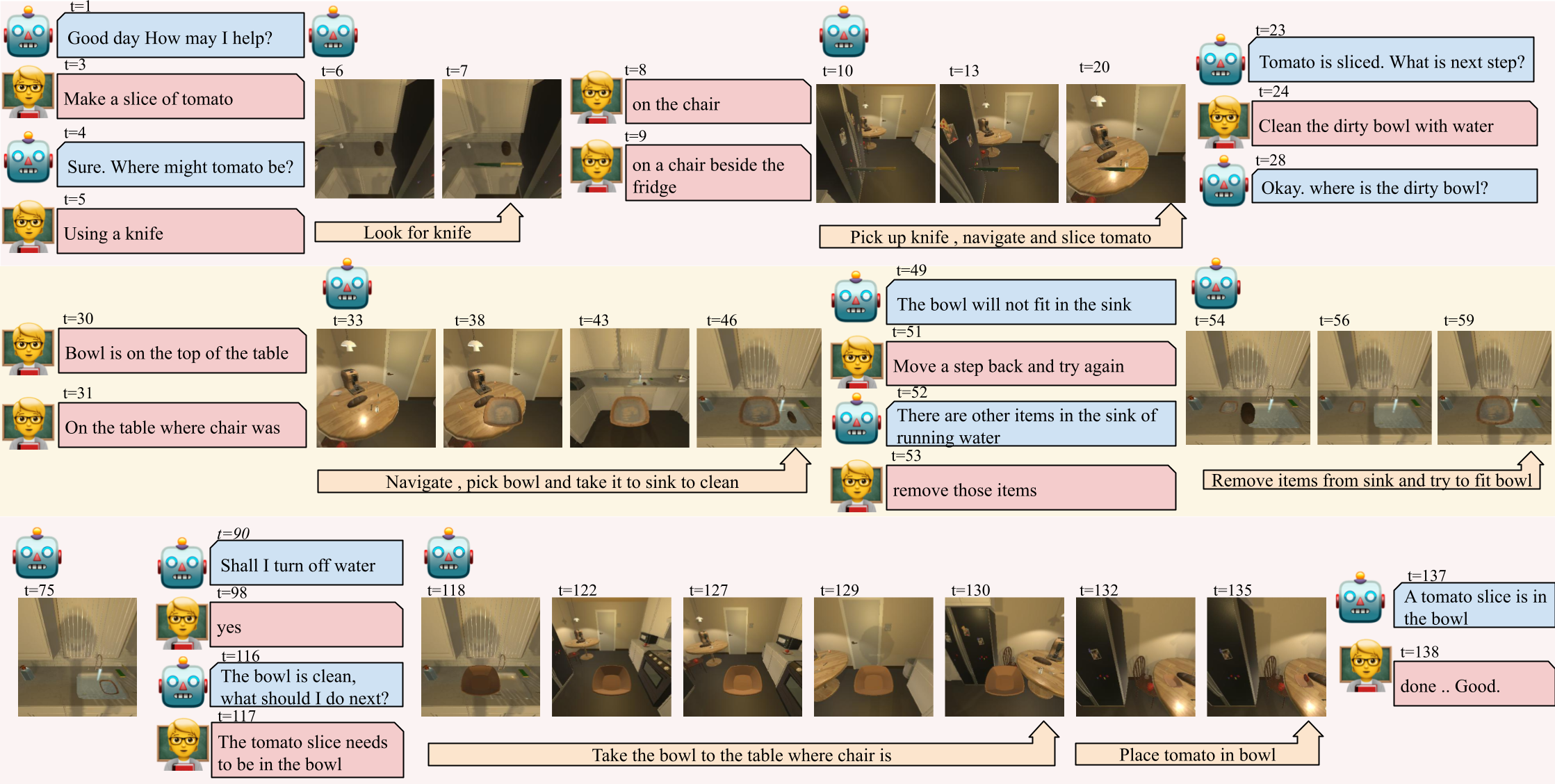}
    \caption{Sample session for the \texttt{N Slices Of X In Y} task. 
    This example demonstrates interleaving chat messages between the \commander\ and \follower\.
    The \commander\ uses referring expressions, such as \textit{On the table where chair was}, to help the \follower\ locate the target object.
    The \follower\ also asks the \commander\ for help, and gives confirmation, frequently.}
    \label{fig:example_n_slices_of_x_in_y}
\end{figure*}

\begin{figure*}[!ht]
    \centering
    \includegraphics[width=1\textwidth]{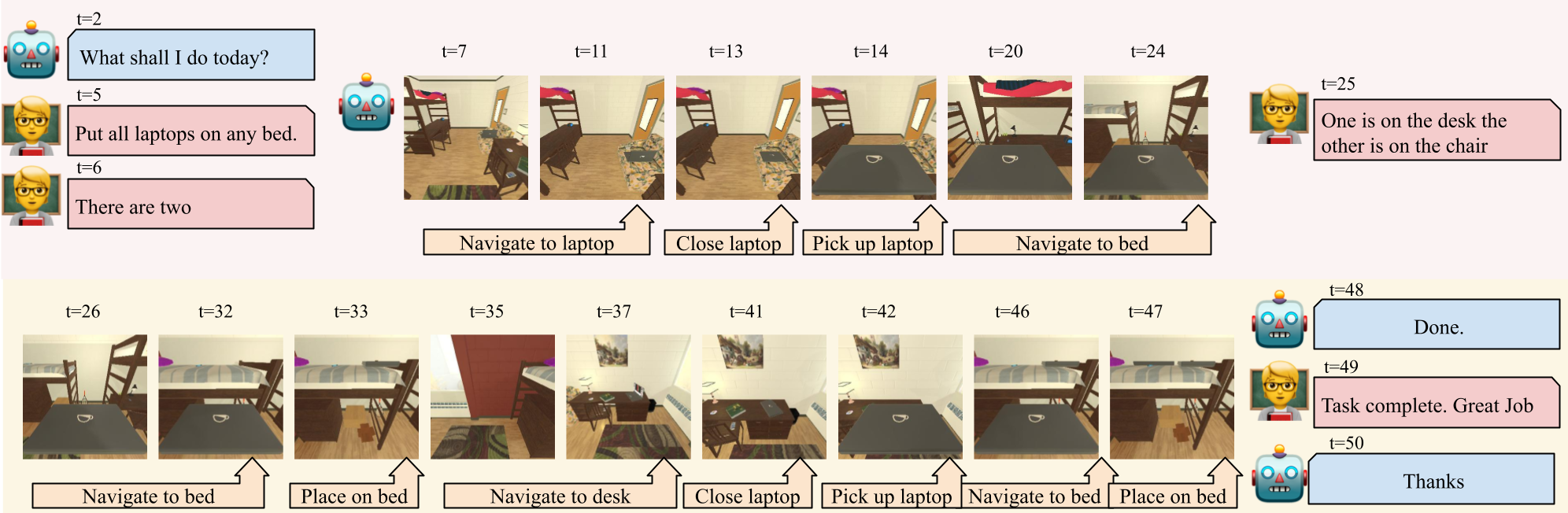}
    \caption{Sample session for the \texttt{Put All X On Y} task in a bedroom.
    The \commander\ intends to give step by step instructions but occasionally provides the next step before the \follower\ has finished the previous step.
    }
    \label{fig:example_put_all_x_on_y}
\end{figure*}

\begin{figure*}[!ht]
    \centering
    \includegraphics[width=1\textwidth]{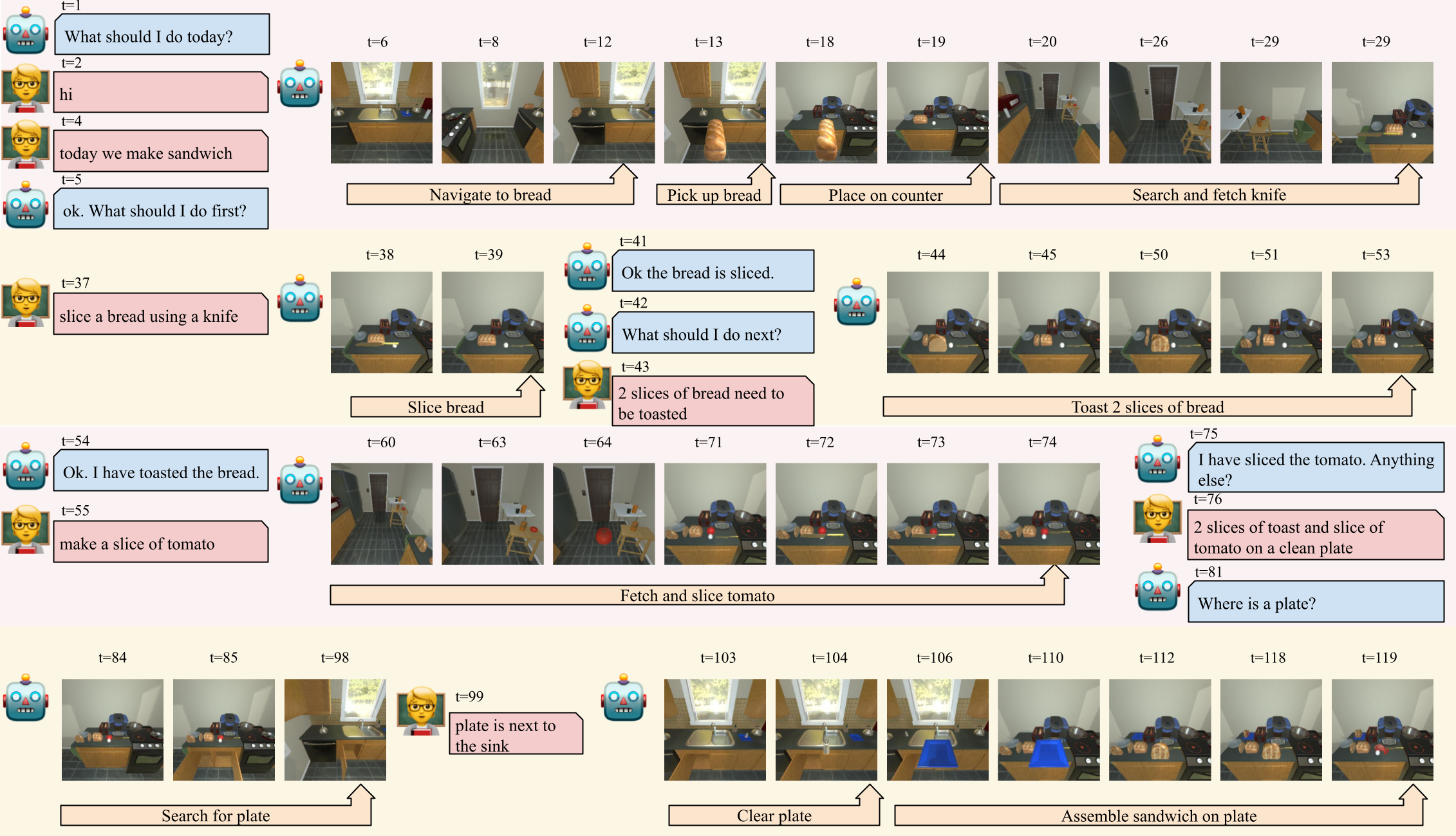}
    \caption{Sample session for the \texttt{Sandwich} task. 
    The \follower\ requires the task of making a sandwich to be broken down into simpler steps but anticipates a few steps, finding the bread and knife before being explicitly asked to.}
    \label{fig:example_sandwich}
\end{figure*}

\begin{figure*}[!ht]
    \centering
    \includegraphics[width=1\textwidth]{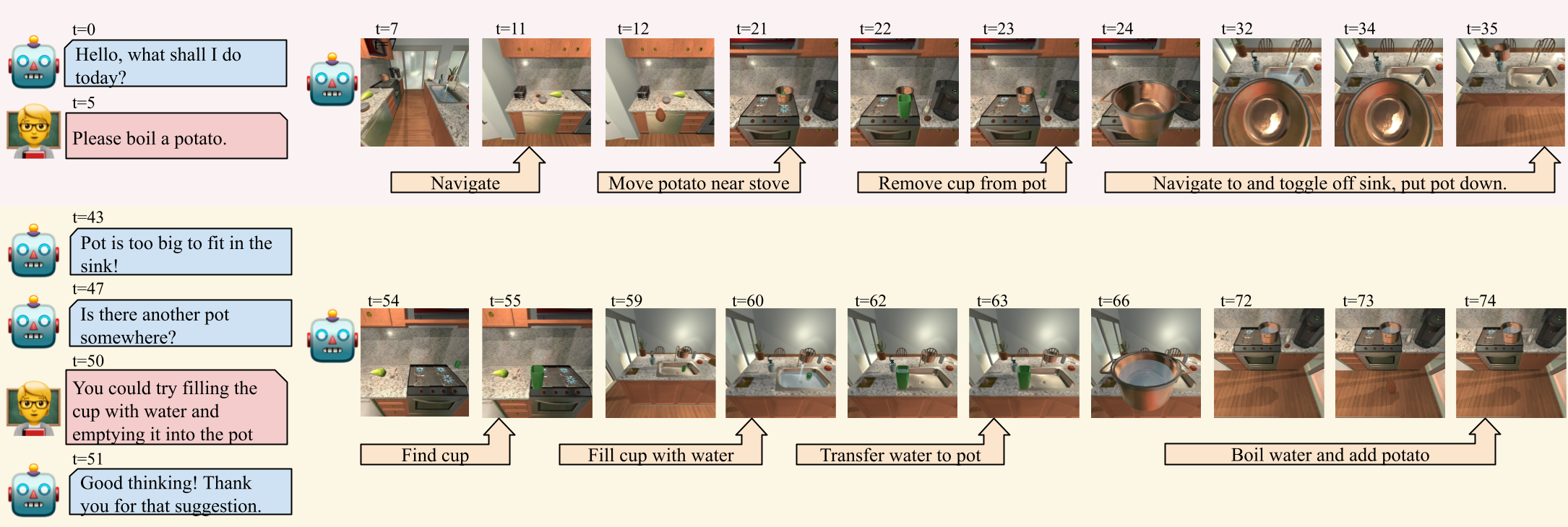}
    \caption{Sample session for the \texttt{Boil Potato} task.
    The session demonstrates an example where the \commander\ helps the \follower\ to solve the issue of a pot not fitting into the sink.
    }
    \label{fig:example_boil_potato}
\end{figure*}

\begin{figure*}[!ht]
    \centering
    \includegraphics[width=1\textwidth]{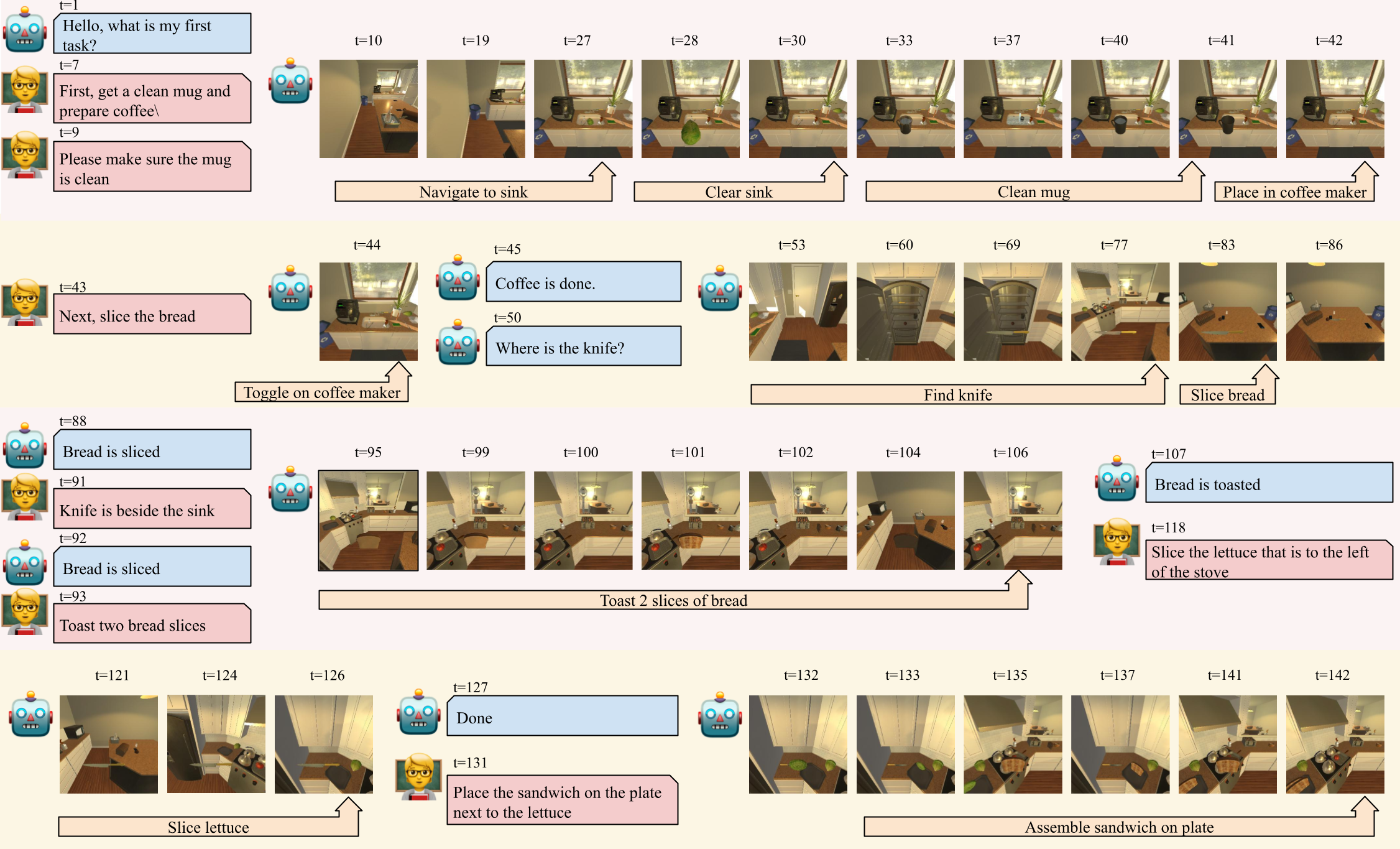}
    \caption{Sample session for the \texttt{Breakfast} task where the \follower\ has to make coffee and a sandwich with lettuce.
    The \commander\ provides step by step instructions but occasionally provides the next step, for example slicing bread, before the \follower\ is done with the previous step, and is sometimes late with help.
    For example, the \follower\ finds the knife alone because the \commander\ does not provide its location.
    }
    \label{fig:example_breakfast}
\end{figure*}

\begin{figure*}[!ht]
    \centering
    \includegraphics[width=1\textwidth]{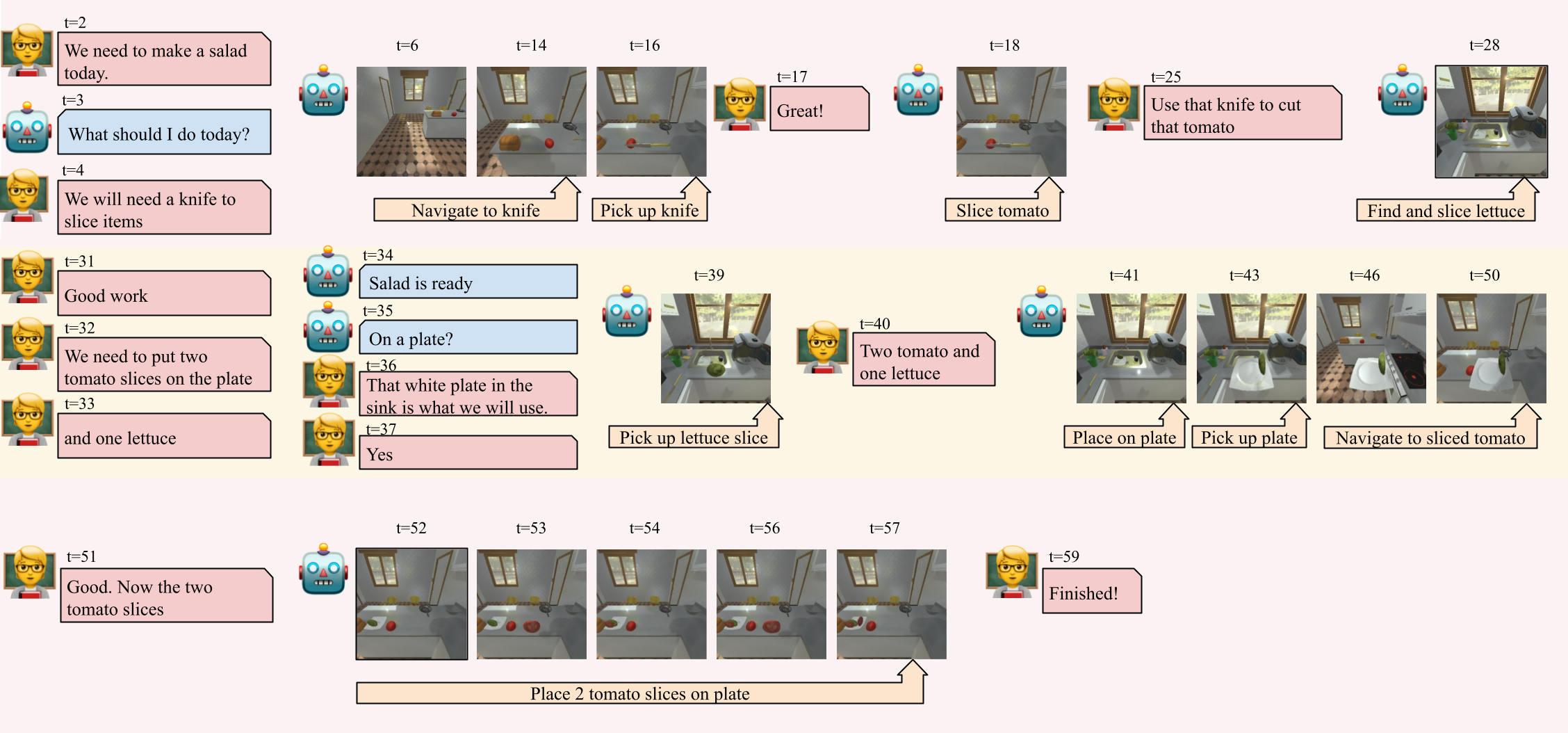}
    \caption{Sample session for the \texttt{Salad} task. 
    The \follower\ anticipates the \commander's directions, slicing the tomato and lettuce before it is asked, but forgets to plate the salad until directed to do so by the \commander.}
    \label{fig:example_salad}
\end{figure*}

\begin{figure*}[!ht]
    \centering
    \includegraphics[width=1\textwidth]{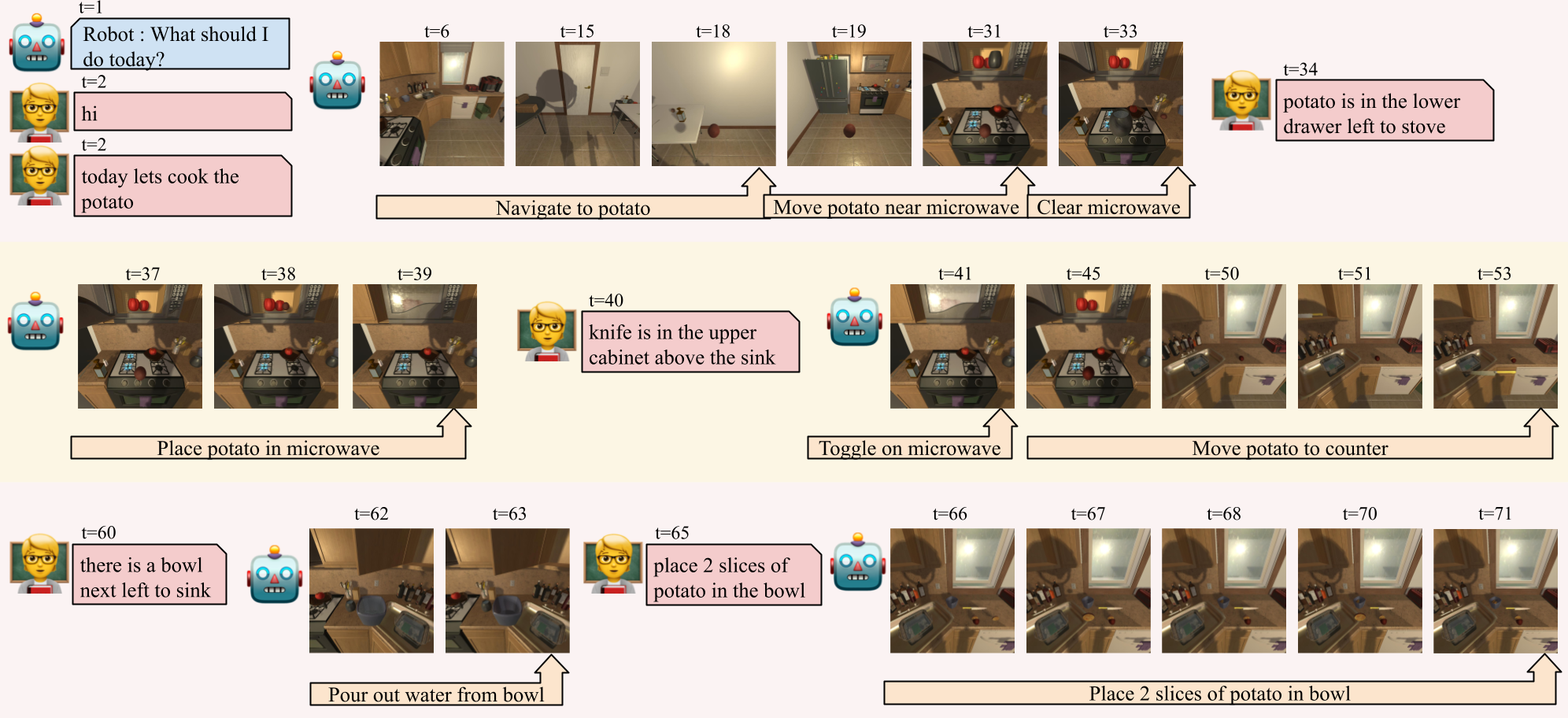}
    \caption{Sample session for the \texttt{N Cooked Slices Of X In Y} task.
    The \follower\ finds a potato before the \commander\ directs it to one.}
    \label{fig:example_n_cooked_slices_of_y}
\end{figure*}

\begin{figure*}[!ht]
    \centering
    \includegraphics[width=1\textwidth]{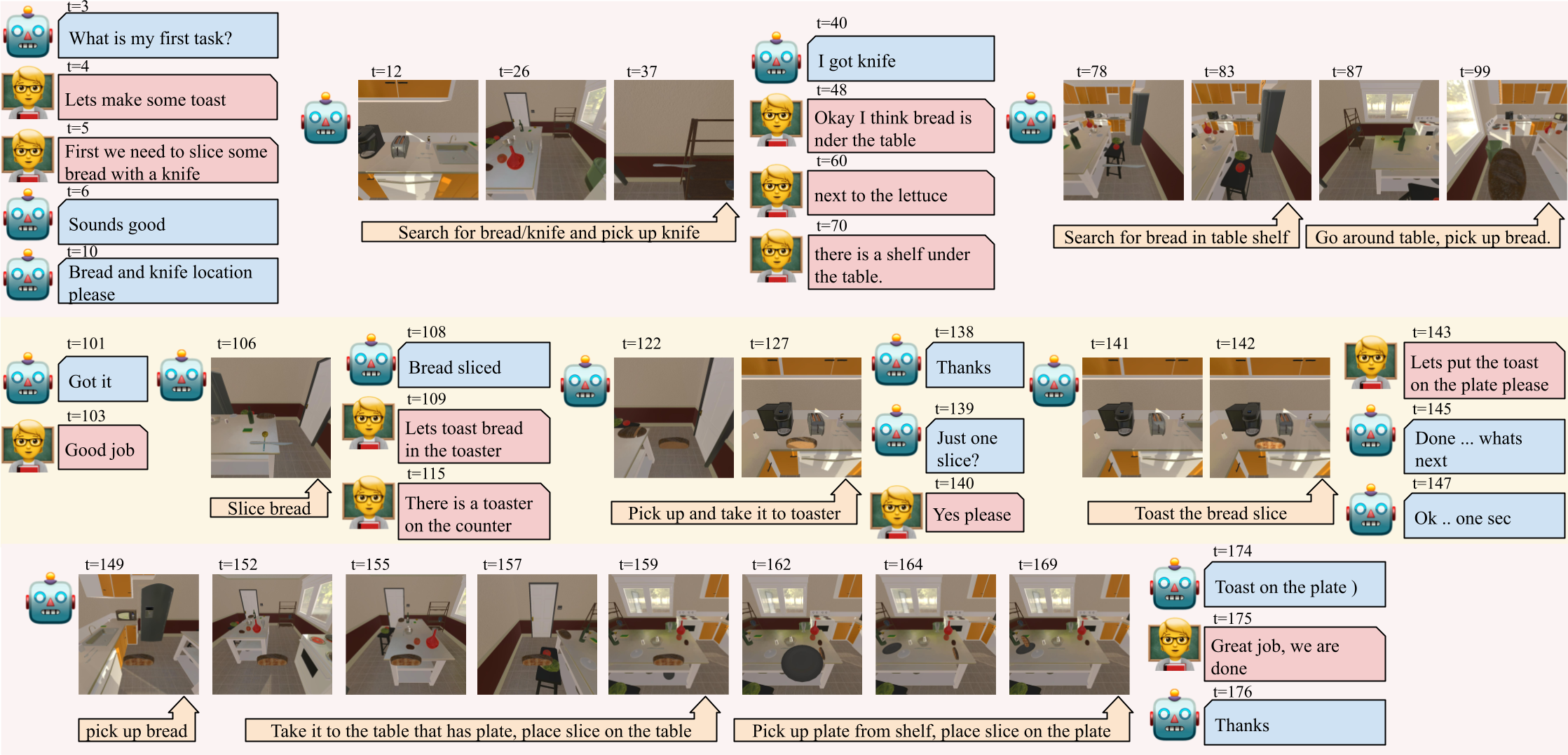}
    \caption{Sample session for the \texttt{Plate Of Toast} task. 
    This session demonstrates interleaving chat messages, referring expressions and \commander\ proving feedback to the \follower\ for sub-tasks.}
    \label{fig:example_plate_of_toast}
\end{figure*}

Consider the example dialogs in Figures \ref{fig:sample_dialog_1} and \ref{fig:Coffee_example_2}. 
In Figure~\ref{fig:sample_dialog_1}, the \commander\ simply tells the \follower\ to prepare coffee, but in Figure~\ref{fig:Coffee_example_2}, the \commander\ provides much lower level instructions, and waits for the \follower\ to complete each step. The initial instruction provided in Figure~\ref{fig:sample_dialog_1} (\textit{``can you make me a coffee please?''}) are similar to goal-level instructions and requests typically seen in task-oriented dialog.
The instructions in Figure~\ref{fig:Coffee_example_2} (\textit{``grab the dirty mug out of the fridge}, \textit{go wash in the sink}) are more similar to the detailed instructions in datasets such as R2R~\cite{anderson:cvpr18}. Every trajectory in ALFRED~\cite{shridhar:cvpr20} is annotated with instructions at both of these levels of granularity.
In \teach, by contrast, dialogues may contain either one or in some cases both levels of granularity.
Thus, \teach\ benchmark agents need to be able to effectively map instructions at different levels of granularity to low level actions.

Dialogues also contain situations where the \commander\ helps the \follower\ get ``unstuck''.
For example, in Figure~\ref{fig:Coffee_example_2}, the \commander\ suggests that the \follower\ needs to clear out the sink in order to place the mug in it. 
In future work, we could attempt to leverage such utterances to learn more general knowledge of the environment that can be used by a \follower\ to get unstuck, either via student forcing from learned rules or by adding hand-written recovery modules analogous to the simple navigation and interaction recovery modules in ABP~\cite{kim:eai21} and EmBERT~\cite{suglia:arxiv21}. 
For example, an agent may use the dialogue in Figure~\ref{fig:Coffee_example_2} to infer that if it tries to place an object in a container and fails, it must try to clear out the container.

In Figure~\ref{fig:Coffee_example_2}, the \follower\ did not explicitly ask for help. 
In contrast, in Figure~\ref{fig:sample_dialog_1}, the \follower\ asks for help when it does not find a mug in the places it initially searches, which prompts the \commander\ to correct their instruction.
This session also illustrates a difference between \teach\, where the task is completed by a human annotator based on online instructions from another human annotator, and benchmarks that elicit descriptions for trajectories generated by a planner, such as ALFRED. 
In Figure~\ref{fig:example_put_all_x_in_one_y}, the \commander\ keeps changing their mind about where the \follower\ should place the tissue boxes, resulting in a less efficient path. 
A human \commander\ may make mistakes when providing instructions, and a human \follower\ may not perfectly follow instructions.
Standard teacher forcing training, including that used in our baseline experiments, does not account for such imperfections in demonstrated trajectories.
However, robust agent models for \teach\ benchmarks will need to learn to identify what information is essential and what is irrelevant or wrong.

Dialogues can contain a lot of feedback, for example the \follower\ informing the \commander\ when it has completed a step or the task, and the \commander\ affirming that a step or task has been completed. 
In the EDH and TfD tasks, an agent will likely need to learn to ignore these feedback steps. 
However, in future, these self-reported completions could be useful to segment large tasks into pragmatic subgoals.
Unlike ALFRED, since our tasks have varying levels of hierarchy, what may constitute pragmatic subgoals for one task may be too much detail for another task.

We place no constraints on our chat interface - for example, we do not impose turn taking. 
Thus, chat messages from the two annotators interleave in interesting ways. 
For example, consider Figure~\ref{fig:example_clean_all_x}. 
The \follower's messages \textit{``What task do I do today?''} and \textit{``I have picked the purple object. What next?''} are preceded by their responses from the \commander. 
An agent performing our EDH or TfD task will need to be able to mentally reorder these messages to successfully complete the task. 
To facilitate detection of interleaved messages, we provide millisecond level timesteps for each action and utterance in the \teach\ data, though in figures we represent each action as a ``timestep.''

The \follower\ can also ask for different kinds of help as they try to complete the task including clarification, for example in Figure~\ref{fig:example_put_all_x_in_one_y}, ``which table? the one with the other tissue box?'', asking for the location of an object, as in Figure~\ref{fig:sample_dialog_1}, \textit{``Where can I find a mug?''}, and help if it is unable to perform an action requested by the \commander, as in Figure~\ref{fig:sample_dialog_1}, \textit{``I can't seem to see a mug''}.
A good \follower\ model should be able to execute actions based on dialogue history, while also being able to interact with the \commander\ in natural language - clarifying ambiguous instructions, obtaining additional information as needed, learning to solve problems, and providing feedback as it completes tasks. 
To accomplish these needs, a model may have to identify different dialog acts, translate dialog history to actions (EDH), detect situations where additional information is needed, and generate appropriate dialog responses in these situations.  
Jointly learning a \commander\ and \follower\ model may begin to enable these strategies.

\end{document}